\documentclass[conference]{IEEEtran}
\usepackage{times}

\usepackage[numbers]{natbib}
\usepackage{multicol}
\usepackage[bookmarks=true]{hyperref}

\usepackage[acronym]{glossaries}
\usepackage{subcaption}
\usepackage{amsmath} 
\usepackage{amssymb} 
\usepackage{gensymb}

\usepackage{wrapfig}
\usepackage{rotating}
\usepackage[export]{adjustbox}
\usepackage{makecell}
\usepackage{multicol}
\usepackage{bm}
\usepackage{wrapfig}

\usepackage[font=small,skip=3pt]{caption}
\glsdisablehyper

\newcommand{\fl}{2.6cm}

\newcommand{\figsize}{2.0cm}
\newcommand{\figsizeL}{2.1cm}
\newcommand{\figsizeT}{3.6 cm}

\newcommand\norm[1]{\left\lVert#1\right\rVert}

\newacronym{wsl}{WSL}{Weakly-Supervised Learning}
\newacronym{fmcw}{FMCW}{Frequency-Modulated Continuous-Wave}
\newacronym{ml}{ML}{Machine Learning}
\newacronym{dl}{DL}{Deep Learning}
\newacronym{cnn}{CNN}{Convolutional Neural Network}
\newacronym{vo}{VO}{Visual Odometry}
\newacronym{ins}{INS}{Inertial Navigation System}
\newacronym{gnss}{GNSS}{Global Navigation Satellite System}
\newacronym{gnssins}{GNSS+INS}{Global Navigation Satellite System and Inertial Navigation System}
\newacronym{rsln}{RSL-Net}{Radar-Satellite Localisation Network}
\newacronym[firstplural=degrees of freedom (DoFs)]{dof}{DoF}{Degree of Freedom}
\newacronym{gan}{GAN}{Generative Adversarial Network}
\newacronym{osm}{OSM}{OpenStreetMap}
\newacronym{api}{API}{Application Programming Interface}
\newacronym{stn}{STN}{Spatial Transformer Net}
\newacronym{pased}{PASED}{Pose-Aware Separable Encoder Decoder}
\newacronym{fov}{FOV}{field-of-view}

\begin{document}

\title{Self-Supervised Localisation between Range Sensors and Overhead Imagery}



\author{\authorblockN{Tim Y. Tang\authorrefmark{1},
Daniele De Martini\authorrefmark{1},
Shangzhe Wu\authorrefmark{2}, and
Paul Newman\authorrefmark{1}}
\authorblockA{\authorrefmark{1}Mobile Robotics Group, University of Oxford \quad  \authorrefmark{2}Visual Geometry Group, University of Oxford}
\{ttang, daniele, szwu, pnewman\}@robots.ox.ac.uk}

\maketitle

\begin{abstract}
Publicly available satellite imagery can be an ubiquitous, cheap, and powerful tool for vehicle localisation when a prior sensor map is unavailable.
However, satellite images are not directly comparable to data from ground range sensors because of their starkly different modalities.
We present a learned metric localisation method that not only handles the modality difference, but is cheap to train, learning in a self-supervised fashion without metrically accurate ground truth.
By evaluating across multiple real-world datasets, we demonstrate the robustness and versatility of our method for various sensor configurations. 
We pay particular attention to the use of millimetre wave radar, which, owing to its complex interaction with the scene and its immunity to weather and lighting, makes for a compelling and valuable use case. 

\end{abstract}

\IEEEpeerreviewmaketitle

\section{Introduction}
The ability to localise relative to an operating environment is central to robot autonomy.
Localisation using range sensors, such as lidars \cite{levinson2010robust,wolcott2015fast} and, more recently, scanning millimetre wave radars \cite{yspark-2019-icra-ws,KidnappedRadarArXiv}, is an established proposition.
Both are immune to changing lighting conditions and directly measure scale, while the latter adds resilience  to weather conditions.

Current approaches to robot localisation typically rely on a prior map built using a sensor configuration that will also be equipped on-board, for example a laser map for laser localisation.
This paper looks at an alternative method.
Public overhead imagery such as satellite images can be a reliable map source, as they are readily available, and often captures information also observable, albeit perhaps in some complex and incomplete way, by sensors on the ground. 
We can pose the localisation problem in a natural way: find the pixel location of a sensor in an overhead (satellite) image given range data taken from the ground. 
The task is, however, non-trivial because of the drastic modality difference between satellite image and sparse, ground-based radar or lidar. 

Recent work on learning to localise a ground scanning radar against satellite images \cite{tang2020rsl} provides a promising direction which addresses the modality difference by first generating a synthetic radar image from a satellite image.
The synthetic image is then ``compared'' against a live radar image for pose estimation.
Such an approach learns metric, cross-modality localisation in an end-to-end fashion, and therefore does not require hand-crafted features limited to a specific environment.

The method in \cite{tang2020rsl} trains a multi-stage network, and needs pixel-wise aligned radar and satellite image pairs for supervision at all stages.
This in turn requires sub-metre accurate ground truth position and sub-degree accurate ground truth heading.
In practice, collecting accurate ground truth requires high-end GPS/\gls{ins}, and possibly bundle adjustment along with other on-board sensor solutions, bringing in burdens in terms of cost and time consumption.

\begin{figure}[!t]
  \centering
  \begin{subfigure}[t]{\fl}
  \includegraphics[width=\fl]{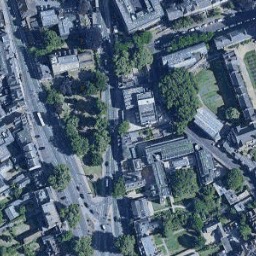}
  \end{subfigure}
  \quad
  \begin{subfigure}[t]{\fl}
  \includegraphics[width=\fl]{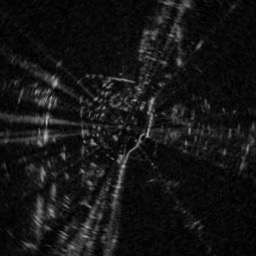}
  \end{subfigure}
\quad
  \begin{subfigure}[t]{\fl}
  \includegraphics[width=\fl]{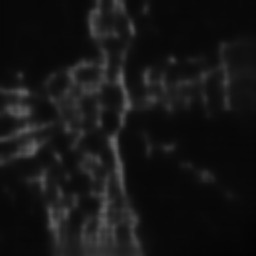}
  \end{subfigure}
  \quad
  \begin{subfigure}[t]{\fl}
  \includegraphics[width=\fl]{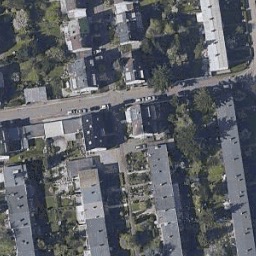}
  \end{subfigure}
 \quad
  \begin{subfigure}[t]{\fl}
  \includegraphics[width=\fl]{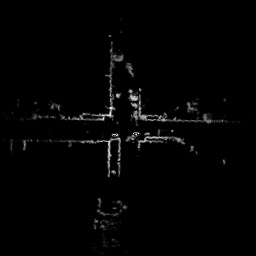}
  \end{subfigure}
  \quad
  \begin{subfigure}[t]{\fl}
  \includegraphics[width=\fl]{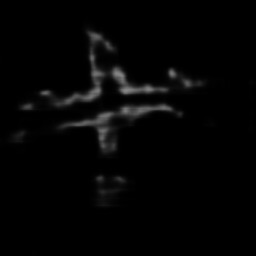}
  \end{subfigure}
  
    \begin{subfigure}[t]{\fl}
  \includegraphics[width=\fl]{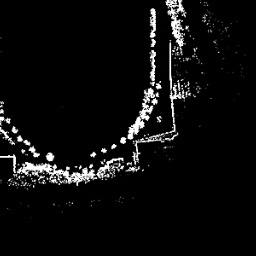}
  \end{subfigure}
 \quad
  \begin{subfigure}[t]{\fl}
  \includegraphics[width=\fl]{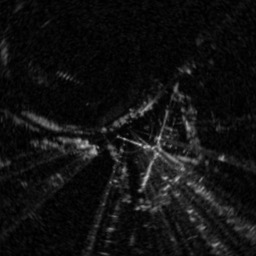}
  \end{subfigure}
  \quad
  \begin{subfigure}[t]{\fl}
  \includegraphics[width=\fl]{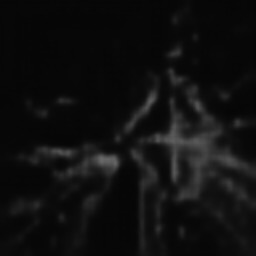}
  \end{subfigure}

  \caption{\label{fig:intro_figs} \footnotesize Given a map image of modality $\mathcal{A}$ (left) and a live data image of modality $\mathcal{B}$ (middle), we wish to find the unknown $SE(2)$ offset between them.
  To do so, our method generates a synthetic image of modality $\mathcal{B}$ (right) that is pixel-wise aligned with the map image, but contains the same appearance and observed scenes as the live data image.
  Top: localising radar data against satellite imagery.
  Middle: localising lidar data against satellite imagery.
  Bottom: localising radar data against prior lidar map.
}%
\vspace{-4mm}
\end{figure}

Building on \cite{tang2020rsl}, we propose a method for localising against satellite imagery that is self-supervised. 
The core idea of both approaches is to generate a synthetic image with the appearance and observed scenes of a live range sensor image, but pixel-wise aligned with the satellite image.
We assume a coarse initial pose estimate is available from place recognition, such that there is reasonable overlap between the live ground sensor field-of-view and a queried satellite image.

Vitally, here we make no use of metrically accurate ground truth for training.
Note also that although designed for localising against satellite imagery, our method can naturally handle other forms of cross-modality registration, such as localising a radar against a prior lidar map.
Figure \ref{fig:intro_figs} shows synthetic images generated by our method used for pose estimation.

To the best of our knowledge, this paper presents the first method to learn the cross-modality, metric localisation of a range sensor in a self-supervised fashion.
Our method is validated experimentally on multiple datasets and achieves performances on-par with a state-of-the-art supervised approach.

\section{Related Work}
\subsection{Localisation Using Overhead Images}
\label{sec:related_work_aerial}
Localisation using aerial or overhead images has been an interest for the community for over a decade.
The methods in \cite{leung2008localization,li2014planar, parsley2010towards} localise a ground camera using aerial images, by detecting Canny edges from aerial imagery, and matching against lines detected by a ground camera.
Several other vision-based approaches project the ground camera images to a top-down perspective via a homography, and compare against the aerial imagery by detecting lane markings \cite{pink2008visual}, SURF features \cite{noda2010vehicle}, or dense matching \cite{senlet2011framework}.
Recent work \cite{chebrolu2019robot} localises a ground robot in a crop field by matching camera features against landmarks from an aerial map, and incorporated semantics of crops to reduce ambiguity.

Metric localisation of range sensors or point-clouds against overhead imagery requires further pre-processing due to the modality difference.
\citet{kaminsky2009alignment} projected point-clouds into images and matched against binary edge images from overhead imagery.
The method in \cite{kaminsky2009alignment} also constructs a ray image by ray-tracing each point, and introduces a free-space cost to aid the image registration.
The work by Veronese et al. \cite{de2015re} accumulates several lidar scans to produce dense lidar intensity images, which are then matched against satellite images utilising Normalised Mutual Information.
Similar as \cite{kaminsky2009alignment}, several other methods also pre-process the aerial image before matching against ground laser observations, for example using edge detection \cite{kummerle2011large} or semantic segmentation \cite{dogruer2010outdoor}.
Our method directly learns the metric localisation of a range sensor end-to-end, without the need for careful pre-processing.

\subsection{Cross-Modality Localisation}
\label{sec:related_work_cml}
Other forms of cross-modality localisation are also heavily studied by the community.
A number of works are proposed to localise a forward facing camera against a prior 3D point-cloud map \cite{wolcott2014visual,caselitz2016monocular,xu20173d}.
\citet{carle2010global} localised a ground laser scanner against an orbital elevation map.
The works in \cite{wangglfp,boniardi2017robust,wang2017flag, mielle2019auto} localise an indoor lidar or stereo camera against architectural floor plans.
\citet{brubaker2013lost} and \citet{floros2013openstreetslam} concurrently proposed matching visual odometry paths to road layouts from OpenStreetMap for localisation.

\subsection{Learning-based State Estimation for Range Sensors}
A number of recent works were proposed for learning odometry or localisation of lidars.
\citet{barsan2018learning} represented lidar data as intensity images, and learned a deep embedding for metric localisation by comparing embeddings of live and map lidar intensity images.
Methods such as \cite{cho2019deeplo,li2019net} learn deep lidar odometry by projecting lidar point-clouds into other representations before passing through the network.
\citet{lu2019l3} used point-clouds as input to learn descriptors, and utilised 3D \glspl{cnn} for solving $SE(2)$ metric localisation by searching in a 3D cost volume.
In their later work, Lu et al. proposed a method to learn $SE(3)$ lidar point-cloud registration end-to-end \cite{Lu_2019_ICCV}.

As an emerging sensor for outdoor state estimation, learning-based methods were proposed for scanning \gls{fmcw} radars.
\citet{aldera2019fast} utilised an encoder-decoder on polar image representation of radar scans to learn key-points for fast classical radar odometry \cite{cen2019radar}.
Barnes et al. learned image-based radar odometry \cite{Barnes2019MaskingByMoving} by masking out regions distracting for pose estimation, and point-based radar odometry \cite{UnderTheRadarICRA2020} by detecting key-points from radar images.
\citet{KidnappedRadarArXiv} encoded images of radar polar scans through a rotation-invariant architecture to perform topological localisation (place recognition).
These methods, however, are designed to compare data of the same sensor type, and does not address modality difference. 
Our approach is similar to \cite{barsan2018learning,cho2019deeplo,aldera2019fast,Barnes2019MaskingByMoving,weston2019probably,KidnappedRadarArXiv,UnderTheRadarICRA2020} in that we also represent lidar and radar data as 2D images prior to passing through the network.

\subsection{Unsupervised Image Generation}
We seek to generate a synthetic image prior to pose computation, where there is no pixel-wise aligned target image for supervision.
CycleGAN \cite{zhu2017unpaired} achieves unsupervised image-to-image transfer between two domains $\mathcal{X}$ and $\mathcal{Y},$ by learning two pairs of generators and discriminators, and enforcing cycle-consistency when an image is mapped from $\mathcal{X}$ to $\mathcal{Y}$ and back from $\mathcal{Y}$ to $\mathcal{X},$ and vice versa.
A number of other methods \cite{liu2017unsupervised,lee2018diverse} also utilise cycle-consistency, but make different assumptions on how the latent spaces of the two domains are treated.
These methods are concerned with generating photo-realistic images.
For the problem of metric localisation, however, we need to explicitly encourage the synthetic image to contain information appropriate for pose estimation.

Several prior works are also geometry-aware.
The methods in \cite{shu2018deforming,wu2019transgaga,xing2019unsupervised} use separate encoders and/or decoders to disentangle geometry and appearance.
The results are networks that can interpolate the geometry and appearance of the output images separately.
Similarly, our method separately encodes information about appearance and the relative pose offset, resulting in an architecture where the two are disentangled.

\section{Overview and Motivation}
We seek to solve for the $SE(2)$ pose between a map image of modality $\mathcal{A}$ and a live data image of modality $\mathcal{B}.$
Our main focus is when modality $\mathcal{A}$ is satellite imagery, while modality $\mathcal{B}$ are range sensor data represented as an image.

Previously, RSL-Net \cite{tang2020rsl} was proposed to solve for the metric localisation between matched pairs of radar and satellite images.
In particular, a synthetic image is generated such that it preserves the appearance and observed scenes of the live radar image, and is pixel-wise aligned with the paired satellite image.
The synthetic image and the live radar image are then projected onto deep embeddings, where their pose offset is found by maximising a correlation surface.
We follow the same general approach, but, unlike RSL-Net, our method learns in a self-supervised fashion.

\subsection{Hand-crafting Features vs. Learning}
A number of works listed in Sections \ref{sec:related_work_cml} and \ref{sec:related_work_aerial} can achieve decent accuracy on localising a ground range sensor against aerial imagery.
However, they typically rely on pre-processing the aerial images using hand-crafted features or transforms designed for a specific set-up and may not generalise to other sensors or different, more complex environments.
For example, \cite{kummerle2011large} focuses on detecting edges from a campus dominated by buildings.
While \cite{de2015re} directly matches accumulated lidar intensity images against aerial imagery without pre-processing, the same method is unlikely to work for radars.

Our data-driven approach instead learns to directly infer the geometric relationship across modalities, remaining free of hand-crafted features.
We show in Section \ref{sec:experimental_validation} when localising against satellite imagery, our method works for various types of scenes including urban, residential, campus, and highway.

\subsection{Generating Images vs. Direct Regression}
A naive approach would be to take a satellite image and a live data image as inputs, and directly regress the pose.
As originally shown in \cite{tang2020rsl}, this led to poor results even for the supervised case.
Our hypothesis is that when the two images are starkly different in appearance and observed scenes, the problem becomes too complex for direct regression to succeed.

Generating synthetic images first prior to pose estimation brings two advantages over directly regressing the pose.
First, generating synthetic images is a simpler and less ill-posed problem than directly regressing the pose, particularly because we can utilise the live data image to condition the generation.
Moreover, to generate images, the network loss is distributed over an entire image of $H \times W$ pixels, where $H$ and $W$ are height and width, instead of on just three parameters ($x, y,$ and $\theta$).
This introduces greater constraint during optimisation.

\subsection{Conditional Image Generation}
Our method of conditional image generation takes in both a map (e.g., satellite) image and a live data image as inputs.
An alternative approach is to learn a domain adaptation from the map modality $\mathcal{A}$ to the live data modality $\mathcal{B},$ without conditioning on the live data image (e.g., standard image-to-image transfer such as CycleGAN \cite{zhu2017unpaired}).

\begin{figure}[!h]
  \centering
  \vspace{-2mm}
  \includegraphics[height=2.4cm]{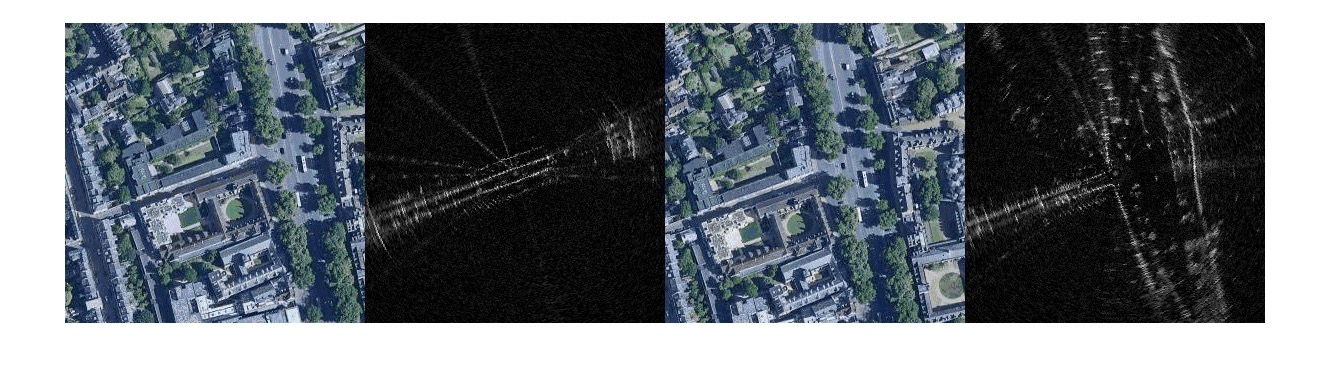}
  \caption{\label{fig:surjective} \footnotesize Two radar images captured 15 seconds apart from each other (2 \& 4), pixel-wise aligned with satellite images (1 \& 3).
Though the overlapping scenes in the satellite images are identical, the radar scans appear significantly different, as they capture different regions in their field-of-view.}%
\end{figure}

In practise, the map (e.g., satellite) image is a denser representation of the environment than a frame of data captured by a range sensor.
Only a fraction of the scenes captured in a satellite map are present in a ground sensor field-of-view, resulting in the scan to appear drastically different depending on the sensor pose.
Shown in Figure \ref{fig:surjective}, the overlapping regions of the two satellite images are identical, while the two radar images observe different regions of the scene.

By using a naive image-to-image transfer approach, there is no guarantee the generated image will contain regions of the scene that are useful for pose comparison against the live data image.
Figure \ref{fig:cyclegan_qual} shows examples of images generated using CycleGAN \cite{zhu2017unpaired}, where the synthetic image highlights different scenes than what are observed by the live data image.
The issue with observability or occlusion can potentially be handled by ray-tracing such as in \cite{kaminsky2009alignment}.
However, not only is this computationally expensive, it does not apply to \gls{fmcw} radars which have multiple range returns per azimuth (in which we are particularly interested).
This problem is inherently addressed by our approach: by conditioning the image generation with the live data image, we can encourage the synthetic image to capture regions of the scene also observed by the live data image, as shown in Sections \ref{sec:methodology} and \ref{sec:experimental_validation}.

\section{Self-Supervised Cross-Modality Localisation}
\label{sec:methodology}
\subsection{Rotation Inference}
Given a paired map (e.g., satellite) image $A \in \mathcal{A}$ and live data image $B \in \mathcal{B}$ with an unknown $SE(2)$ offset, we seek to generate a synthetic image that contains the same appearance and observed scenes as $B,$ but is pixel-wise aligned with $A.$

Let the $SE(2)$ pose difference between $A$ and $B$ be parametrised as $\begin{bmatrix}
x & y & \theta
\end{bmatrix}^T,$ such that by rotating $B$ by $\theta,$ followed by a translation of $\begin{bmatrix}
x & y 
\end{bmatrix}^T,$ one can pixel-wise align $B$ onto $A.$
The image generation can be formulated as:
\begin{equation}
\label{eqn:naive_generation}
f(A, B) \rightarrow \tilde{B}_{\theta, \alpha}
\end{equation}
where $\alpha  = \begin{bmatrix}
x & y
\end{bmatrix}^T.$
$\tilde{B}_{\theta, \alpha}$ is a generated image of modality $\mathcal{B}$ that synthesises the input live sensor image $B$ applied with a rotation of $\theta, $ followed by a translation of $\alpha  = \begin{bmatrix}
x & y
\end{bmatrix}^T.$
Thus, $\tilde{B}_{\theta, \alpha}$ is pixel-wise aligned with the input map image $A,$ but contains the same observed scenes as $B.$

However, as originally noted in \cite{tang2020rsl}, the mapping in \eqref{eqn:naive_generation} is difficult to learn as the inputs $A$ and $B$ are offset by both a translation and a rotation.
\glspl{cnn} are inherently \textit{equivariant} to translation, but not to rotation \cite{lenc2015understanding}.
As a result, the \glspl{cnn} in the generator cannot automatically utilise their mutual information and thereby capture their geometric relationship.

The method in \cite{tang2020rsl} proposes to infer the rotation prior to image generation.
Namely, reducing \eqref{eqn:naive_generation} to two steps:
\begin{equation}
\label{eqn:rotation}
f_R(A, B) \rightarrow B_{\theta}
\end{equation}
\vspace{-4mm}
\begin{equation}
\label{eqn:generation}
f_G(A, B_{\theta}) \rightarrow \tilde{B}_{\theta, \alpha}
\end{equation}

Here $f_R$ is a function that infers the rotation offset $\theta$ between $A$ and $B,$ and outputs $B_\theta,$ which is input image $B$ rotated by $\theta.$
Now, $B_\theta$ is rotation-aligned with the map frame, and therefore offset with $A$ only by a translation, which \glspl{cnn} can naturally handle.
$f_G$ is an image generation function that produces the synthetic image $\tilde{B}_{\theta, \alpha}.$
The experiments in \cite{tang2020rsl} show that learning \eqref{eqn:rotation} and \eqref{eqn:generation} sequentially resulted in better performance than learning \eqref{eqn:naive_generation} directly, as the former is congruous with the equivariance properties of \glspl{cnn}.

\begin{figure}[!h]
  \centering
  \includegraphics[height=5.5cm]{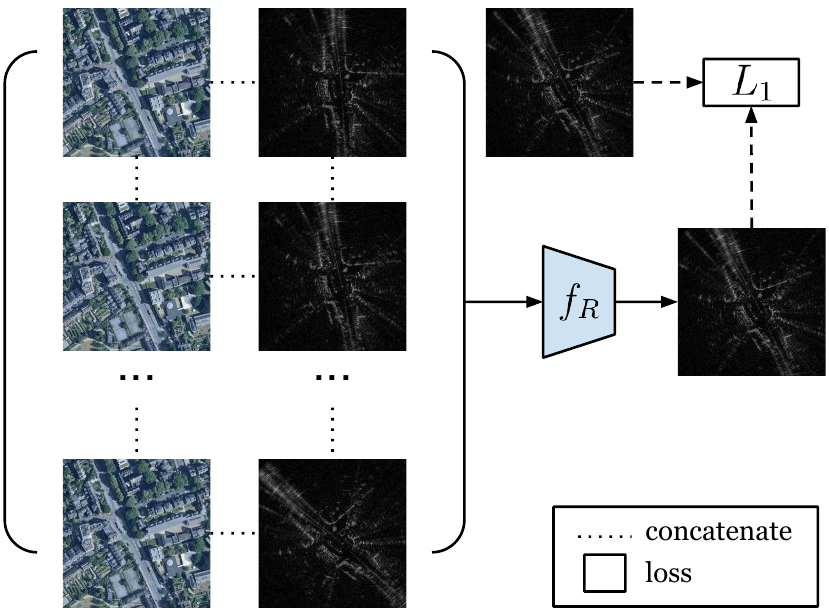}
  \caption{\label{fig:rotation_network_supervised} \footnotesize Prior work in \cite{tang2020rsl} proposes a network to infer the rotation offset.
  The rotation offset is found by softmaxing a stack of rotated radar images to produce a radar image with the same heading as the satellite image.}%
\end{figure}

In \cite{tang2020rsl}, the rotation inference function $f_R$ is parametrised by a deep network as shown in Figure \ref{fig:rotation_network_supervised}, where satellite imagery and radar images are used as an example.
Given a coarse initial heading estimate, the live data image $B$ is rotated a number of times with small increments to form a stack of rotated images $\{ B \} =  \{B_{\theta_0}, B_{\theta_1}, \dots, B_{\theta_n}\}$, where the number of rotations $n$ and the increment are design parameters.
Each rotated image is further concatenated with the map image to form a stacked tensor input of $n$ pairs of map and live data images.
The output of the network is a softmaxed image from $\{B\}$ that corresponds to $B$ rotated to be rotation-aligned with $A,$ namely $B_\theta.$
The core idea is that the network $f_R$ will assign a large softmax weight to the image from $\{B\}$ whose heading most closely aligns with the map image $A,$ and small weights to all other images in $\{ B \}$.

If metrically accurate heading ground truth $\theta$ is available, then one can rotate $B$ to form an image target to $B_\theta$ used for supervising the rotation inference, as in Figure \ref{fig:rotation_network_supervised}.
In this work we assume this is never the case, thus the network for $f_R$ must learn to infer the rotation offset self-supervised.

\begin{figure*}[!htbp]
\centering
\includegraphics[width=0.86\textwidth]{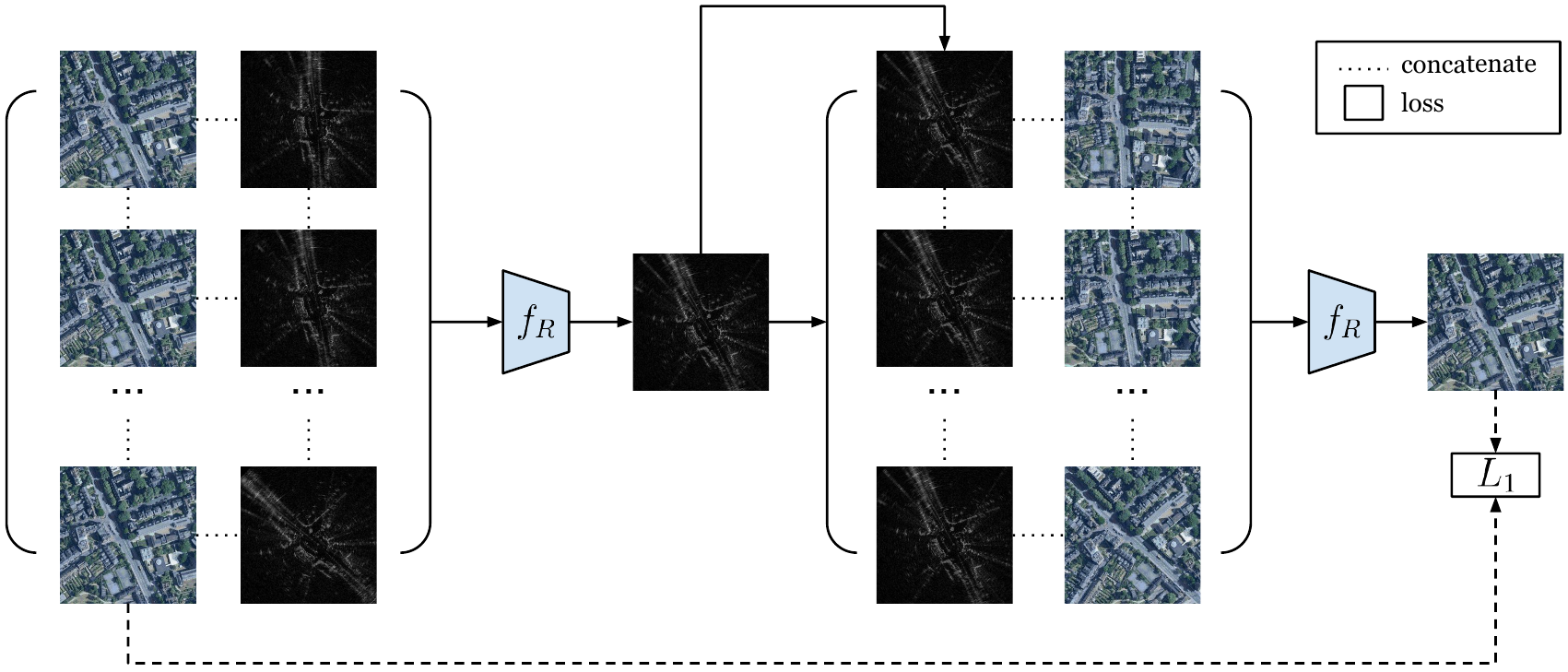}
\caption{\footnotesize \label{fig:rotation_ss} Given $A$ and a rotation stack $\{ B \}$ the network $f_R$ finds $B_{\theta_i}$ by taking softmax.
Then, given $B_{\theta_i}$ and a rotation stack $\{ A \},$ the network outputs a softmaxed map image from  $\{ A \}.$
 A loss is applied to enforce the output of the second pass to be $A,$ which in turn enforces the output of the first pass to be $B_\theta.$
 Here both symbols for $f_R$ in the figure refer to the same network, but at different forward passes. }
\end{figure*}

For this reason, while following the same architecture as \cite{tang2020rsl}, our method for inferring rotation uses a different training strategy that enables self-supervised learning.
In order for the network $f_R$ to produce the correct output, it must be able to infer the rotation from the solution space $\{ B \},$ despite there being a modality difference between map image $A$ and live data image $B.$
We make the observation that if the network can infer the rotation offset from a stack of rotated live data images $\{ B \}$, then, given a live data image $B_{\theta_i}$, $f_R$ should also be able to output $A_{\theta_i}$ from a stack of rotated map images $\{ A \},$ where $A_{\theta_i}$ is rotation-aligned with $B_{\theta_i}.$
Specifically, if we have $B_{\theta_i} = B_\theta,$ then the softmaxed map image from $\{ A \}$ should be $A,$ as $A$ and $B_\theta$ are rotation-aligned.

As such, to learn rotation inference self-supervised, we need to pass through the network $f_R$ twice.
The first pass is identical as in the supervised approach in Figure \ref{fig:rotation_network_supervised}, where we denote the output softmaxed image as $B_{\theta_i}.$
$B_{\theta_i}$ is then used as input to the second pass through network $f_R,$ together with
a stack of map images $\{ A \} = \{A, A_{\phi_0}, A_{\phi_1}, \dots, A_{\phi_m} \}.$ 
The rotation angles $ \begin{bmatrix}
\phi_0 & \phi_1 &\dots & \phi_m
\end{bmatrix}$ can be chosen randomly, and the order of $\{ A \}$ is shuffled such that the original non-rotated map image $A$ can be at any index within $\{ A \}.$
Each image is concatenated with $B_{\theta_i}$ to form the input stack for passing through $f_R$ the second time.
The network is supervised with an $L_1$ loss that enforces the output of the second pass to be the non-rotated map image $A,$ which in turn enforces the output of the first pass $B_{\theta_i}$ to be $B_\theta,$ as $B_\theta$ is rotation-aligned with $A.$
Our approach is shown in Figure \ref{fig:rotation_ss}.
We use an increment of $2 \degree$ when forming the rotation stack $\{ B \}.$

The estimate for the rotation offset, $\hat{\theta},$ can then be found from the arg-softmax for the rotation stack $\{ B \}.$ 

\subsection{Image Generation}
\label{sec:image_generation}
\begin{figure}[!h]
  \centering
  \includegraphics[height=2.6cm]{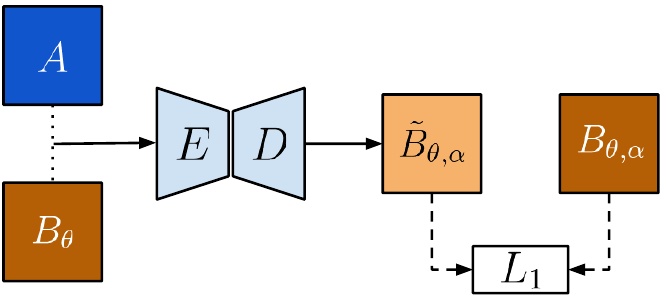}
  \caption{\label{fig:generator_network_supervised} \footnotesize Architecture for image generation in prior supervised approach \cite{tang2020rsl}.}%
\vspace{-2mm}
\end{figure}

Given $A$ and $B_\theta$ we seek to generate a synthetic image $\tilde{B}_{\theta, \alpha}$ as in \eqref{eqn:generation}, where $\tilde{B}_{\theta, \alpha}$ is pixel-wise aligned with $A.$
\cite{tang2020rsl} learns the image generation function by a supervised approach, concatenating $A$ and $B_\theta,$ and applying an encoder-decoder architecture, as shown in Figure \ref{fig:generator_network_supervised}.
This is possible since a target for the synthetic image $\tilde{B}_{\theta, \alpha}$ can be obtained by applying the ground truth transform.

\begin{figure*}[!htbp]
\centering
\includegraphics[width=0.85\textwidth]{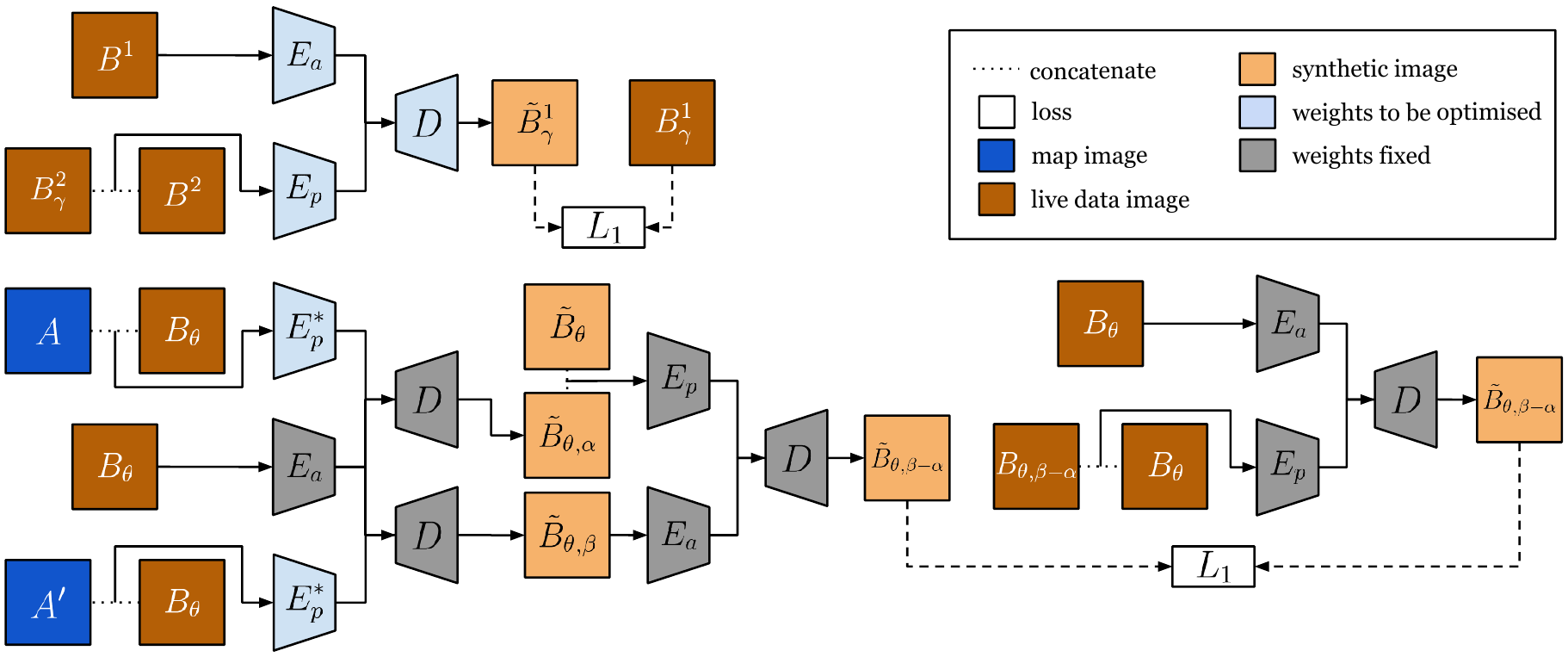}
\caption{\footnotesize \label{fig:generator_network_ss} Top: during pre-training, we can learn an appearance encoder $E_a,$ and a pose encoder $E_p$ that discovers the translation offset between an image of $\mathcal{B}$ and a shifted version of itself.
Bottom: Taking $E_a,$ $E_p,$ and $D$ and fixing their weights, we seek to learn $E_p^*$ which discovers the translation offset between two images from different modalities.
$E_a,$ $E_p,$ and $D$ can provide the necessary geometric and appearance relationships used for learning $E_p^*$ self-supervised.
 }
\end{figure*}

To generate synthetic images self-supervised, we propose an architecture we call \gls{pased}, shown in Figure \ref{fig:generator_network_ss}.
\gls{pased} is trained in two steps: the first is a pre-training, intra-modality process that can be supervised (top half of Figure \ref{fig:generator_network_ss}), while the second handles cross-modality comparison (bottom half of Figure \ref{fig:generator_network_ss}).

Taking two random images $B^1$ and $B^2$ in the live data modality $\mathcal{B}$ from the training set, where $B^1$ and $B^2$ can be at arbitrary heading, we apply a known translation offset $\gamma \in \mathbb{R}^2$ to $B^2.$
This forms an image $B^2_\gamma$ that is a shifted version of $B^2.$
We pass $B^1$ through an appearance encoder $E_a$ that encodes its appearance and observed scenes.
$B^2_\gamma$ and $B^2$ are passed as inputs to a pose encoder $E_p$ that encodes the translation offset between the input images.
The latent spaces from $E_a$ and $E_p$ are combined before passing through a decoder $D,$ which outputs a synthetic image $\tilde{B}^1_\gamma$ that is $B^1$ shifted by a translation $\gamma.$
In other words, \gls{pased} discovers the translation offset between the two images passed as input to $E_p,$ and applies the latent translation encoding to the input image of $E_a.$
The pre-training can be supervised as $\gamma$ is known, thus we can shift $B^1$ by $\gamma$ to produce the target $B^1_\gamma.$
The fact that we use different images $B^1$ and $B^2$ for inputs to $E_a$ and $E_p$ ensures appearance and pose are disentangled from each other.
As shown later, this allows modules of \gls{pased} to be separated and re-combined with newly learned modules.

In the second step, we fix the weights of $E_a,$ $E_p,$ and $D$ which are optimised from the pre-training step. 
This narrows down the self-supervision problem to learning a cross-modality pose encoder $E_p^*$ that discovers the translation offset between an image of modality $\mathcal{A}$ and another of $\mathcal{B}.$
Taking $A$ and $B_\theta$ as inputs, $E_p^*$ should encode the unknown translation offset $\alpha$ between them.
Concurrently, $B_\theta$ is encoded by $E_a,$  where the latent space is combined with the latent space produced by $E_p^*(A, B_\theta),$ before decoded by $D.$
This encoder-decoder combination will generate a synthetic image $\tilde{B}_{\theta, \alpha},$ which we do not have a target for.

We can apply a known shift to the centre position of $A$ to query another map image $A',$ where $A'$ is offset with $B_\theta$ by an unknown translation $\beta.$
Using the same encoder-decoder combination as before, we can take $A'$ and $B_\theta$ to generate a synthetic image $\tilde{B}_{\theta, \beta}.$
Furthermore, given $B_\theta$ and the networks learned from pre-training, we can easily generate $\tilde{B}_{\theta}$ by encoding a zero shift.
If we pass $\tilde{B}_\theta$ and $\tilde{B}_{\theta, \alpha}$ to the pre-trained pose encoder $E_p,$ then the latent space will encode a shift of $-\alpha.$
Combing this latent space with $E_a(\tilde{B}_{\theta, \beta}),$ we can decode a synthetic image $\tilde{B}_{\theta, \beta-\alpha}.$
Here $\beta - \alpha$ is a known value as it is the translation offset applied to $A$ to obtain $A'.$

We can shift $B_\theta$ by $\beta-\alpha$ to get $B_{\theta, \beta - \alpha}.$
Using $B_{\theta, \beta - \alpha}$ and $B_\theta,$ we can generate $\tilde{B}_{\theta, \beta - \alpha}$ with $E_a,$ $E_p,$ and $D,$ shown on the bottom right of Figure \ref{fig:generator_network_ss}.
A loss can then be established between the two synthetic images  $\tilde{B}_{\theta, \beta - \alpha},$ where the latter one is a target image created by modules with weights fixed.
By back-propagation the loss optimises the network $E_p^*$.
Alternatively we can use $B_{\theta, \beta - \alpha}$ as the target, but using $\tilde{B}_{\theta, \beta - \alpha}$ led to faster convergence.

For the loss to be minimised, two conditions must hold true.
First, $\tilde{B}_{\theta, \beta}$ must have correctly encoded the appearance and observed scenes in $B_\theta.$
Second, $\tilde{B}_{\theta, \alpha}$ and $\tilde{B}_{\theta, \beta}$ must have the correct translations $\alpha$ and $\beta,$ respectively.
By satisfying these two constraints we can ensure $E_p^*$ is able to discover the translation offset across modalities, and is compatible with pre-trained networks $E_a$ and $D$ for image generation.

\begin{figure}[!h]
  \centering
  \includegraphics[width=0.42\textwidth]{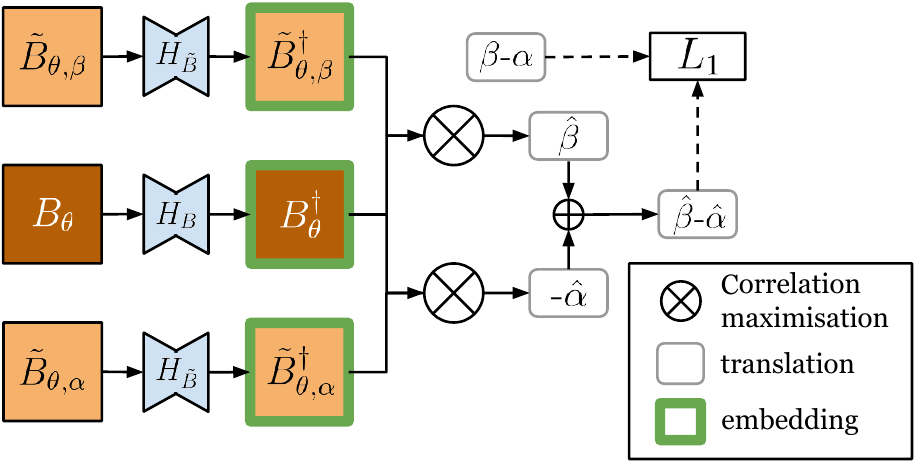}
  \caption{\label{fig:embedding_ss} \footnotesize The networks $H_B$ and $H_{\tilde{B}}$ are learned to project real live images and synthetic images to a joint embedding, where their translation offset can be found by maximising correlation.}%
\vspace{-3mm}
\end{figure}

\begin{figure*}[!ht]
\centering
\includegraphics[width=0.9\textwidth]{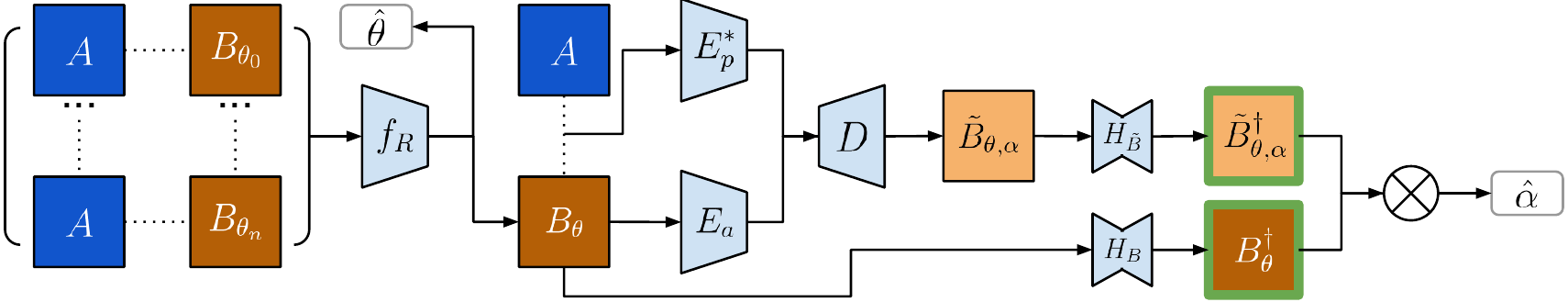}
\caption{\footnotesize \label{fig:inference_network} Overall data flow of our method at inference: given map image $A$ and live data image $B,$ based on the initial heading estimate, we form a stack of rotated images $\{B_{\theta_0}, \dots, B_{\theta_n}\},$ from which $f_R$ discovers $B_\theta$ that is $B$ rotated to be rotation-aligned with $A.$
This process also infers the heading estimate $\hat{\theta}.$
$A$ and $B_\theta$ are used to generate a synthetic image $\tilde{B}_{\theta, \alpha}$ that has the same appearance and observed scene as $B_\theta$ and is pose-aligned with $A.$
$\tilde{B}_{\theta, \alpha}$ and $B_\theta$ are projected to deep embeddings $\tilde{B}_{\theta, \alpha}^\dagger$ and $B_\theta^\dagger,$ where the estimate for the translation offset $\hat{\alpha}$ is found by correlation maximisation.}
\end{figure*}

\subsection{Pose Estimation}
Taking $\tilde{B}_{\theta, \alpha}$ and $B_\theta,$ we embed them to a joint space, where their translation offset is found by maximising correlation on the learned embeddings.
This can be performed efficiently in the Fourier domain, as is done in prior works that use a similar approach \cite{barsan2018learning,Barnes2019MaskingByMoving,tang2020rsl}.
In this step, we can infer $\hat{\alpha} = \begin{bmatrix}
\hat{x} & \hat{y}
\end{bmatrix}^T,$ which is our posterior estimate to the translation.

The embeddings are thus learned to further ensure the synthetic image and the live image can be correctly correlated.
Without ground truth $\alpha,$ we can self-supervise using a similar approach as in learning \gls{pased}, by applying a known shift.
The architecture for learning the embeddings is shown in Figure \ref{fig:embedding_ss}, where we denote the embedding network for real and synthetic images to be $H_B$ and $H_{\tilde{B}},$ respectively.
Given learned deep embeddings $\tilde{B}_{\theta, \beta} ^\dagger$ and $B_\theta^\dagger,$ the translation offset by correlation maximisation is found to be $\hat{\beta}.$
If we replace $\tilde{B}_{\theta, \beta} ^\dagger$ with $\tilde{B}_{\theta, \alpha} ^\dagger$ and reverse the order, the offset found will be $-\hat{\alpha}.$
The sum of the two offsets $\beta - \alpha$ is known, and can be used to establish a loss term.
Similar as in Section \ref{sec:image_generation}, $\tilde{B}_{\theta, \beta}$ is obtained by shifting the map image $A$ to get $A'.$

\section{Experimental Validation}
\label{sec:experimental_validation}
The overall pipeline for data flow at inference time is shown in Figure \ref{fig:inference_network}.
The inference runs at about $10~\mathrm{Hz}$ on a single 1080 Ti GPU.
We evaluate on a large number of public, real-world datasets collected with vehicles equipped with on-board range sensors.
The datasets we use come with metric ground truths that are decently accurate, though we noticed the GPS/\gls{ins} solutions in certain places can drift up to a few metres.
 
We add large artificial pose offsets to the ground truth when querying for a satellite image, thereby simulating a realistic robot navigation scenario where the initial pose estimate can solve place recognition, but is too coarse for the robot's metric pose.
Using a map (e.g., satellite) image queried at this coarse initial pose estimate, our method solves metric localisation by comparing against the live sensor data.
The true pose offsets are hidden during training as our method is self-supervised, and are only revealed at test time for evaluation purposes.

The artificial offset is chosen such that the initial estimate has an unknown heading error in the range $[-\frac{\pi}{8}, \frac{\pi}{8}],$ therefore given the initial estimate $\theta_0,$ the rotation inference must choose a solution space of at least $[\theta_0 - \frac{\pi}{8}, \theta_0 + \frac{\pi}{8}]$ to guarantee the correct solution can be found.
We use a pixel-wise translation error in the range $[-25, 25]$ pixels.
Depending on the resolution for a specific experiment, this corresponds to an error of at least $[-10\textrm{m}, 10\textrm{m}]$ and up to more than $[-20\textrm{m}, 20\textrm{m}].$

\subsection{Radar Localisation Against Satellite Imagery}
\label{sec:rsl_experiment}
We evaluate on two datasets with \gls{fmcw} radar and GPS: the Oxford Radar RobotCar Dataset \cite{RadarRobotCarDatasetICRA2020} and the MulRan Dataset \cite{gskim-2020-mulran}.
The satellite images for RobotCar are queried using Google Maps Platform \cite{google}.
For MulRan they are queried using Bing Maps Platform \cite{bing}, as high-definition Google satellite imagery is unavailable at the place of interest.

\begin{figure}[!t]
  \centering
  \begin{subfigure}[t]{\figsizeT}
  \includegraphics[width=\figsizeT]{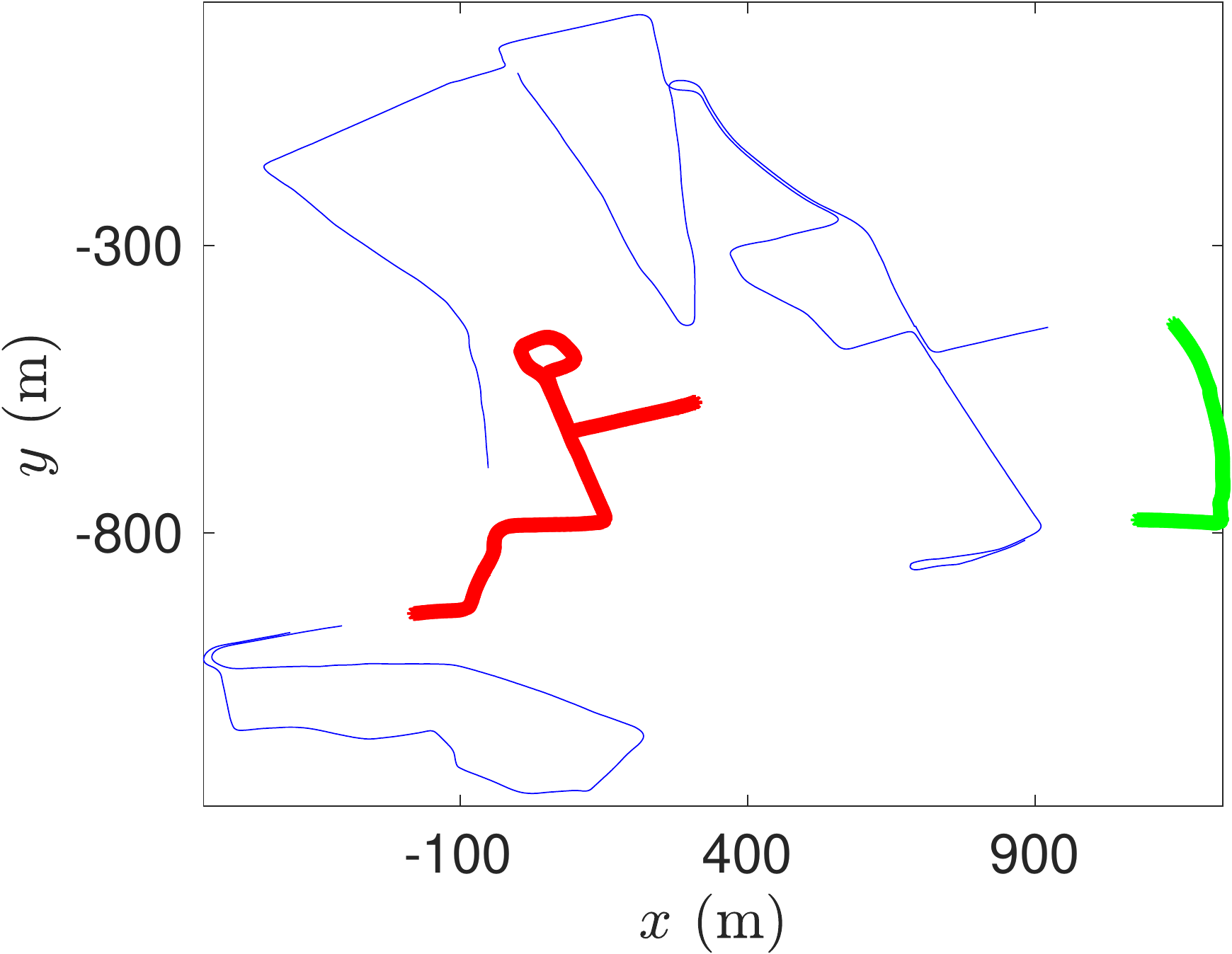}
  \end{subfigure}
\quad
  \begin{subfigure}[t]{\figsizeT}
  \includegraphics[width=\figsizeT]{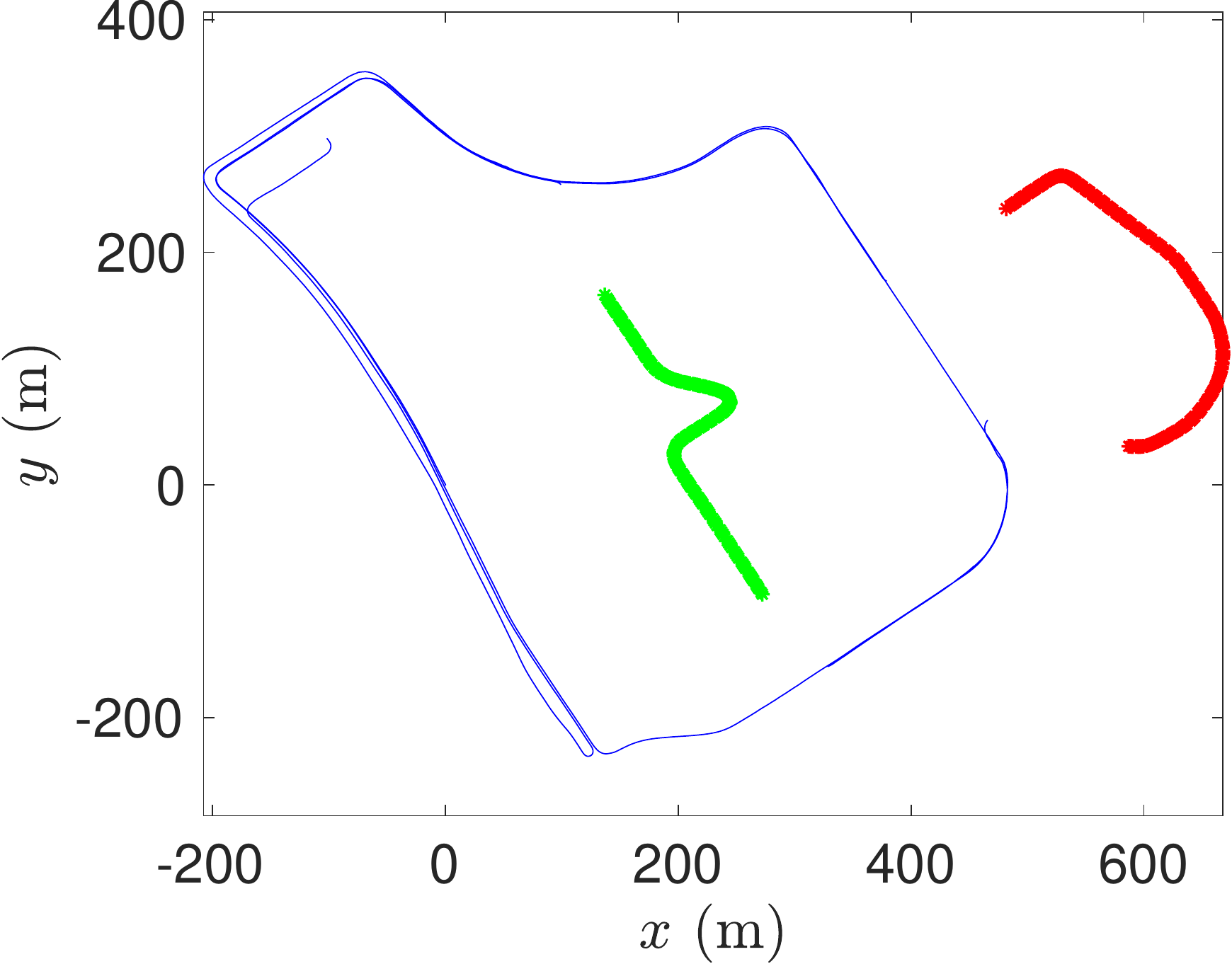}
  \end{subfigure}

  \begin{subfigure}[t]{\figsizeT}
  \includegraphics[width=\figsizeT]{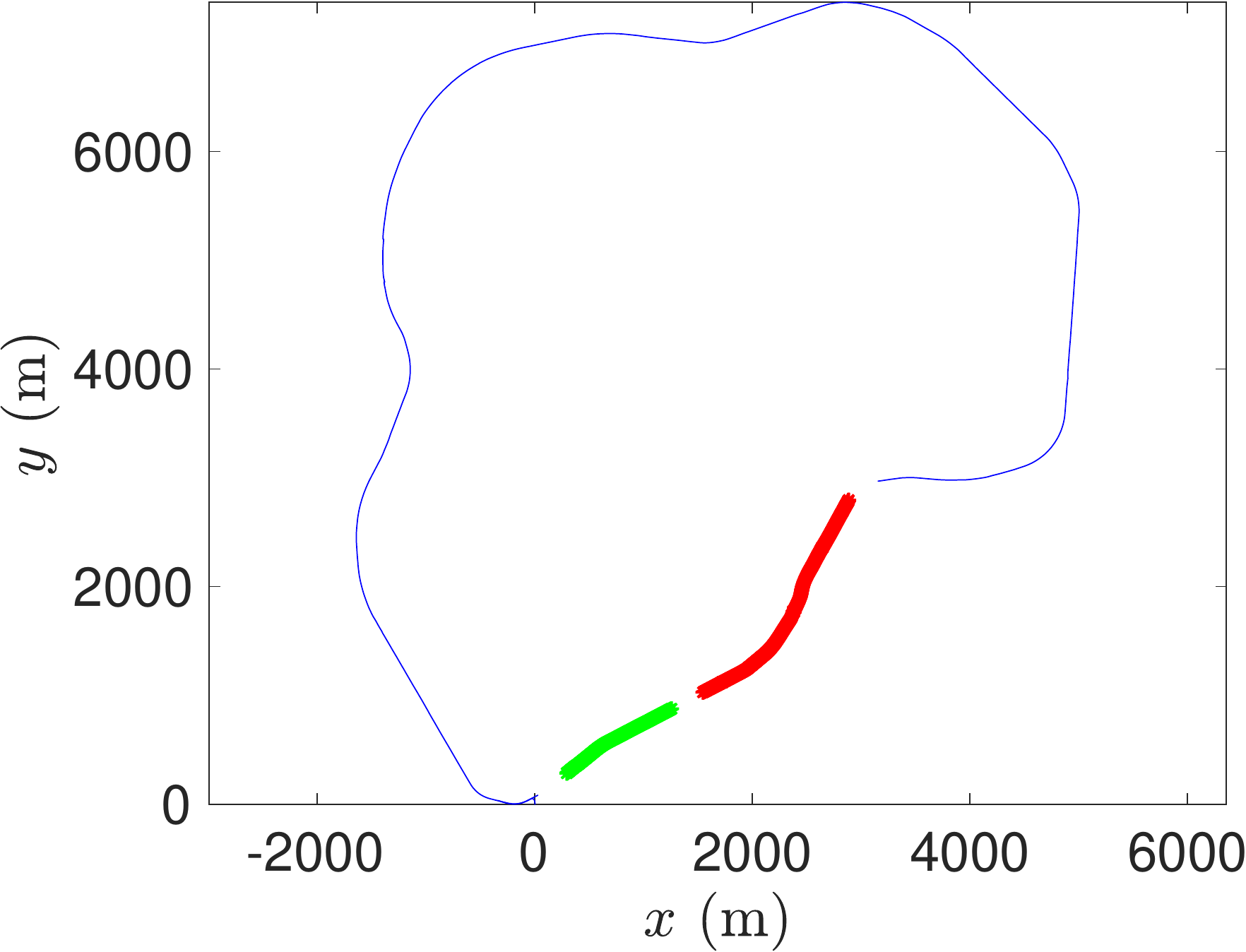}
  \end{subfigure}
\quad
  \begin{subfigure}[t]{\figsizeT}
  \includegraphics[width=\figsizeT]{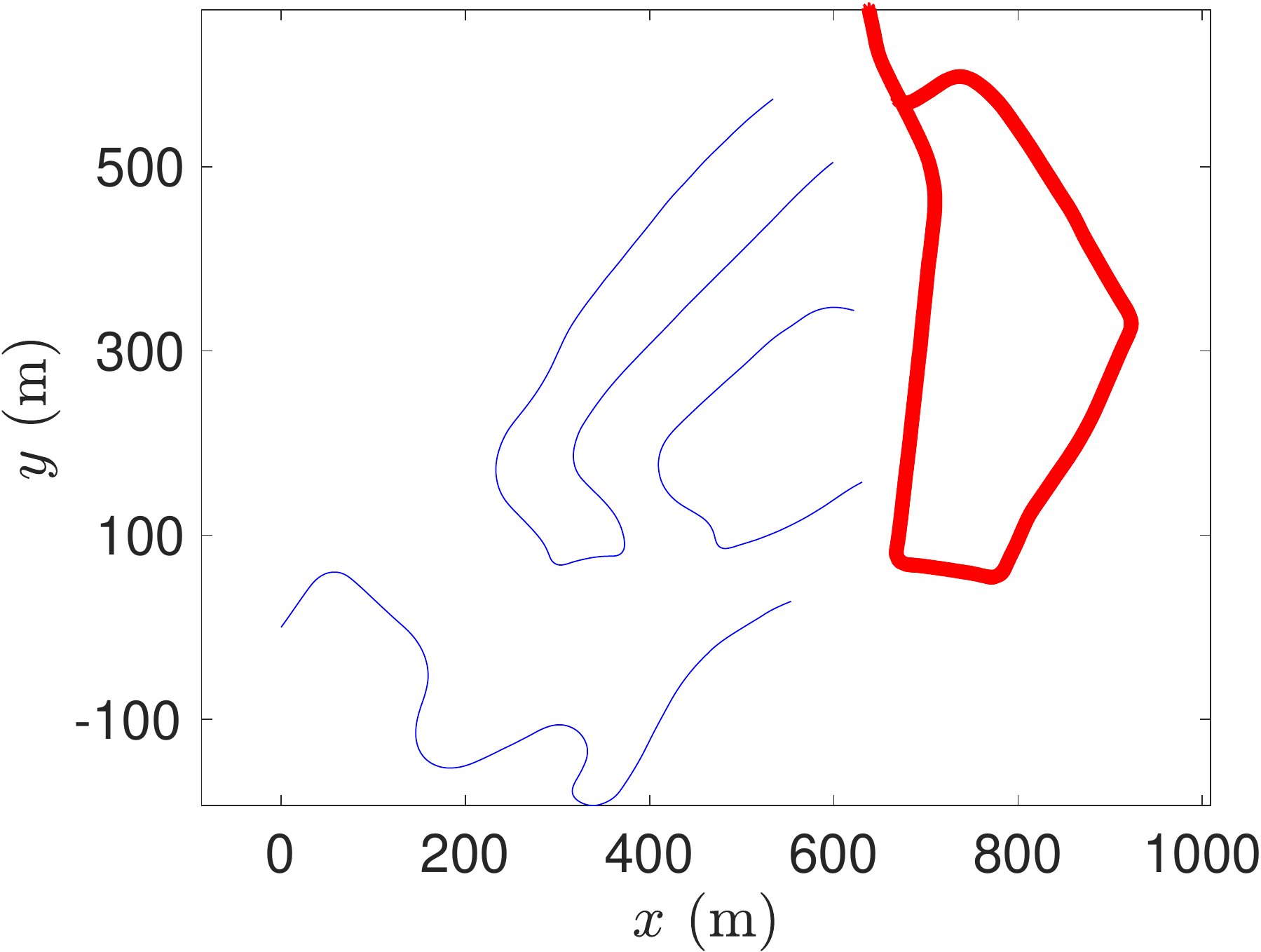}
  \end{subfigure}

  \caption{\label{fig:splits} \footnotesize Training (blue), validation (green), and test (red) trajectories for RobotCar (top left), \texttt{KAIST} (top right), \texttt{Sejong} (bottom left) and \texttt{20111003\_drive0034} (bottom right).
  Certain data are removed to avoid overlap between the splits.
}%
\end{figure}

We benchmark against the prior supervised method RSL-Net \cite{tang2020rsl} in our experiments, which is evaluated only on the RobotCar Dataset.
Both datasets contain repeated traversals of the same routes.
We separately train, validate, and test for every dataset, splitting the data as in Figure \ref{fig:splits}.
For the RobotCar Dataset, we split the trajectories the same way as in \cite{tang2020rsl} for a fair comparison.
For the RobotCar Dataset, the training set consists of training data from sequences no. 2, no. 5, and no. 6, while we test on the test data from sequence no. 2.
For the MulRan Dataset, we used sequences \texttt{KAIST 01} and \texttt{Sejong 01}.
The RobotCar test set features an urban environment, while \texttt{KAIST 01} is in a campus and \texttt{Sejong 01} is primarily a highway.

\begin{table}[!h]
\begin{tabular}{c|ccc|cc}
\hline
                & \multicolumn{3}{c|}{\textbf{Mean Error (metric)}} & \multicolumn{2}{c}{\textbf{(pixel)}} \\
                & $x~\mathrm{(m)}$               & $y~\mathrm{(m)}$              & $\theta~\mathrm{(\degree)}$               & $x$                      & $y$               \\ \hline
RobotCar (ours) &       3.44      &    5.40      &   3.03               &             3.97              &           6.23              \\
MulRan (ours)   &       6.02     &     7.02     &       2.92      &         7.64               &                 8.91   \\   \hline
\begin{tabular}[c]{@{}c@{}}RobotCar \\ (RSL-Net \cite{tang2020rsl}, supervised)\end{tabular} &       2.74      &   4.26      &   3.12              &             3.16              &          4.92              \\
MulRan (RSL-Net)   &        5.85     &     7.11     &       1.88      &         7.42               &                 9.03
\end{tabular}
\captionof{table}{\footnotesize \label{tab:results_rs} 
Mean error for radar localisation against satellite imagery.}
\end{table}

We test on every fifth frame, resulting in 201 frames from the RobotCar Dataset and 358 from the MulRan Dataset, spanning a total distance of near $4~\mathrm{km}.$
The resolution used is $0.8665~\textrm{m/pixel}$ for RobotCar and $0.7876~\textrm{m/pixel}$ for MulRan.
The mean errors are reported in Table \ref{tab:results_rs}.

\subsection{Lidar Localisation Against Satellite Imagery}
\label{sec:lsl_experiment}
For this experiment, we evaluate on the RobotCar Dataset  \cite{RadarRobotCarDatasetICRA2020} which also has two Velodyne HDL-32E lidars mounted in a tilted configuration, and KITTI (raw dataset) \cite{Geiger2013IJRR} which has a Velodyne HDL-64E lidar and GPS data.

For the RobotCar Dataset, the trajectories are split into training, validation, and test sets approximately the same way as in Section \ref{sec:rsl_experiment}.
For the KITTI Dataset, the training set includes sequences \texttt{20110929\_drive0071}, \texttt{20110930\_drive0028}, and  \texttt{20111003\_drive0027}.
Sequence \texttt{20110926\_drive0117} is used for validation.
Finally, data in \texttt{20111003\_drive0034} are split into training and test, as shown in Figure \ref{fig:splits}.
To turn 3D lidar point-clouds to lidar images, the point-clouds are projected to the $x-y$ plane.
We discard points with $z$ values smaller than zero to remove ground points when creating the lidar images.

Since lidars have a shorter range than radars, we use satellite images of a greater zoom level, with resolution $0.4332~\textrm{m/pixel}$ for RobotCar and $0.4592~\textrm{m/pixel}$ for KITTI.
The test set consists of 200 frames for RobotCar and 253 for KITTI, spanning a total distance of near $3~\mathrm{km}.$
The test set for KITTI features a residential area.
The results are reported in Table \ref{tab:results_ls}.

\begin{table}[!h]
\begin{tabular}{c|ccc|cc}
\hline
                & \multicolumn{3}{c|}{\textbf{Mean Error (metric)}} & \multicolumn{2}{c}{\textbf{(pixel)}} \\
                & $x~\mathrm{(m)}$               & $y~\mathrm{(m)}$              & $\theta~\mathrm{(\degree)}$               & $x$                      & $y$               \\ \hline
RobotCar (ours) &       1.54      &    1.85      &   2.34               &             3.55             &           4.27              \\
KITTI (ours)   &       3.05     &     3.13    &       1.67      &         6.64               &            6.82  \\   \hline
RobotCar (RSL-Net)  &       2.31      &    2.55      &   2.08              &             5.33              &          5.89             \\
KITTI (RSL-Net)   &         2.45     &     2.79     &       1.59      &         5.34              &                6.08
\end{tabular}
\captionof{table}{\footnotesize \label{tab:results_ls} 
Mean error for lidar localisation against satellite imagery.}
\vspace{-2mm}
\end{table}

\subsection{Radar Localisation Against Prior Lidar Map}
\vspace{-2mm}
\begin{figure*}[!htbp]
\vspace{2mm}
\centering
\begin{minipage}{.6\textwidth}
  \adjincludegraphics[height=4cm,clip]{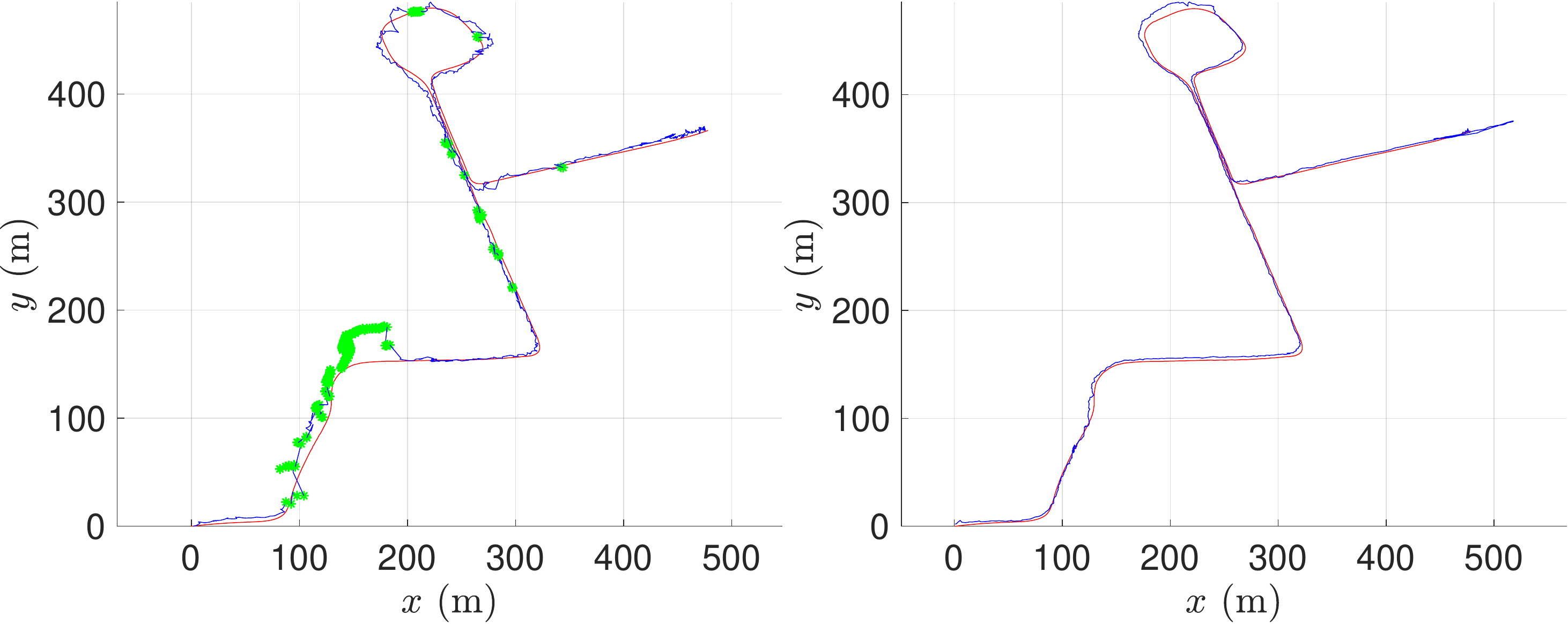}
  \captionsetup{justification=raggedright}
\caption{\label{fig:systems_result} \footnotesize Estimated pose (blue) vs. ground truth pose (red) for localising a radar (left) and \\ a lidar (right) against satellite imagery.
Our system continuously tracks the vehicle's pose over $1~\mathrm{km}$, where we occasionally fall back to odometry for the radar experiment (green).
Our system is stand-alone and requires GPS only for the first frame.
}
\end{minipage}%
\begin{minipage}{.35\textwidth}
\vspace{-4mm}
    \adjincludegraphics[height=4.6cm,clip]{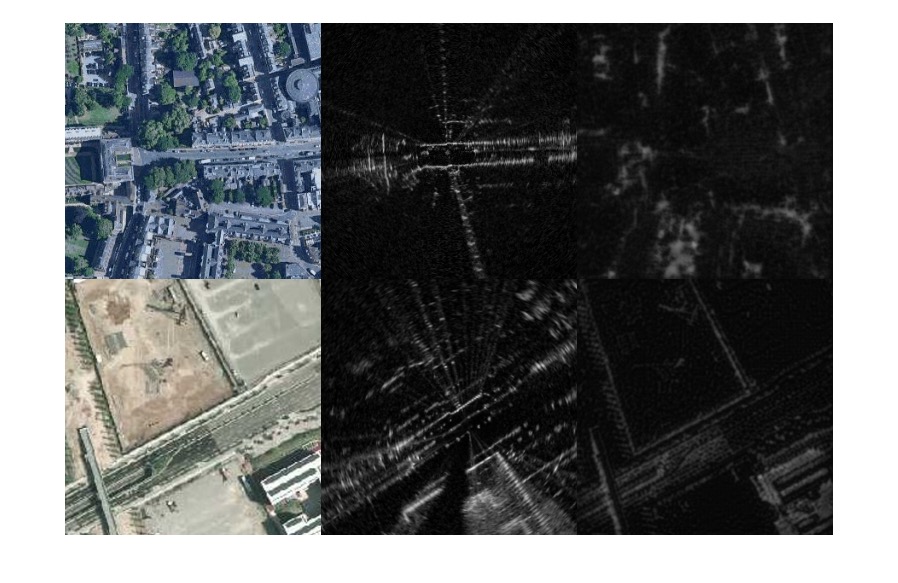}
    \captionof{figure}{\footnotesize\label{fig:cyclegan_qual} Results of CycleGAN: satellite image $A$ (left), ground truth radar image $B_{\theta, \alpha}$ (middle), synthetic radar image $\tilde{B}_{\theta, \alpha}$ (right).
    This led to large localisation error as $\tilde{B}_{\theta, \alpha}$ does not contain scenes observed by $B.$}
\end{minipage}
\vspace{-4mm}
\end{figure*}

\begin{figure*}[h]%
    \centering
    \begin{subfigure}{\figsizeL}
    \centering
    \begin{subfigure}{\figsize}{\includegraphics[width=\figsize]{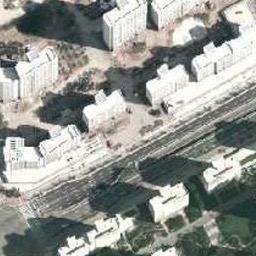}}\end{subfigure}
    \begin{subfigure}{\figsize}{\includegraphics[width=\figsize]{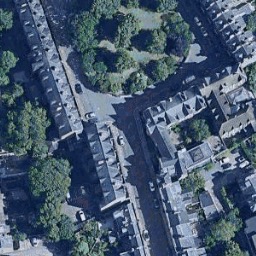}}\end{subfigure}
    \begin{subfigure}{\figsize}{\includegraphics[width=\figsize]{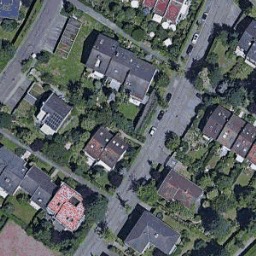}}\end{subfigure}
    \begin{subfigure}{\figsize}{\includegraphics[width=\figsize]{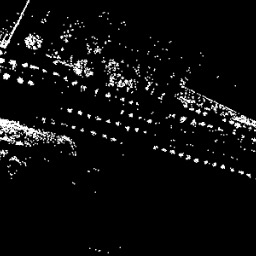}}\end{subfigure}
    \caption{}
    \end{subfigure}
    \begin{subfigure}{\figsizeL}
    \centering
    \begin{subfigure}{\figsize}{\includegraphics[width=\figsize]{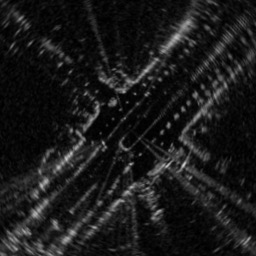}}\end{subfigure}
    \begin{subfigure}{\figsize}{\includegraphics[width=\figsize]{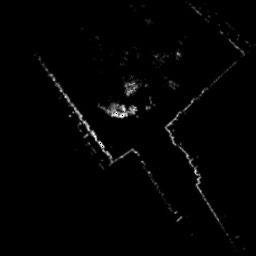}}\end{subfigure}
    \begin{subfigure}{\figsize}{\includegraphics[width=\figsize]{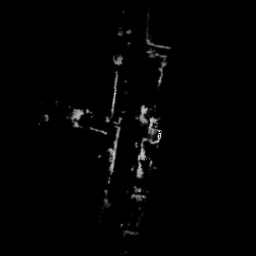}}\end{subfigure}
    \begin{subfigure}{\figsize}{\includegraphics[width=\figsize]{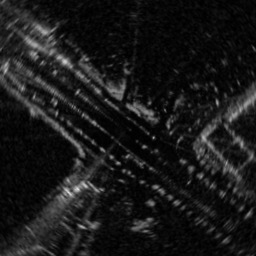}}\end{subfigure}
    \caption{}
    \end{subfigure}
    \begin{subfigure}{\figsizeL}
    \centering
    \begin{subfigure}{\figsize}{\includegraphics[width=\figsize]{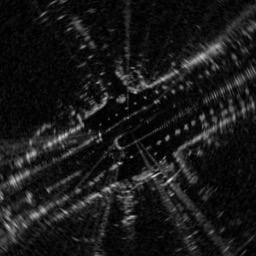}}\end{subfigure}
    \begin{subfigure}{\figsize}{\includegraphics[width=\figsize]{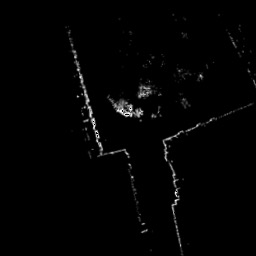}}\end{subfigure}
    \begin{subfigure}{\figsize}{\includegraphics[width=\figsize]{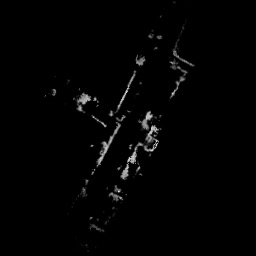}}\end{subfigure}
    \begin{subfigure}{\figsize}{\includegraphics[width=\figsize]{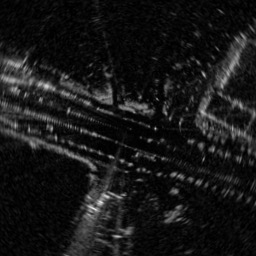}}\end{subfigure}
    \caption{}
    \end{subfigure}
    \begin{subfigure}{\figsizeL}
    \centering
    \begin{subfigure}{\figsize}{\includegraphics[width=\figsize]{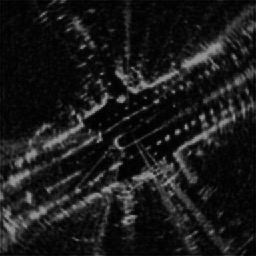}}\end{subfigure}
    \begin{subfigure}{\figsize}{\includegraphics[width=\figsize]{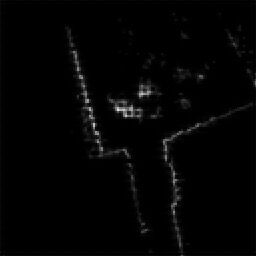}}\end{subfigure}
    \begin{subfigure}{\figsize}{\includegraphics[width=\figsize]{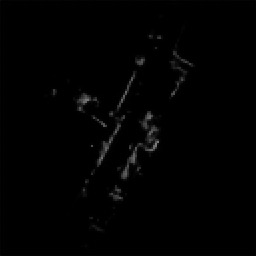}}\end{subfigure}
    \begin{subfigure}{\figsize}{\includegraphics[width=\figsize]{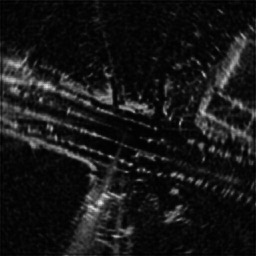}}\end{subfigure}
    \caption{}
    \end{subfigure}
    \begin{subfigure}{\figsizeL}
    \centering
    \begin{subfigure}{\figsize}{\includegraphics[width=\figsize]{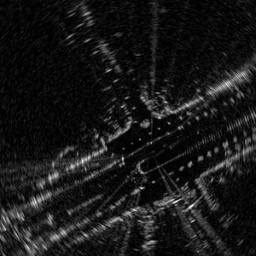}}\end{subfigure}
    \begin{subfigure}{\figsize}{\includegraphics[width=\figsize]{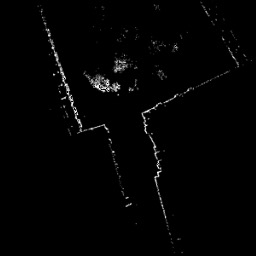}}\end{subfigure}
    \begin{subfigure}{\figsize}{\includegraphics[width=\figsize]{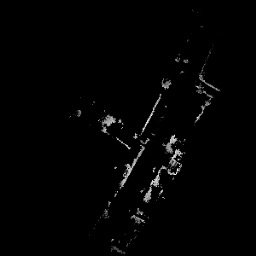}}\end{subfigure}
    \begin{subfigure}{\figsize}{\includegraphics[width=\figsize]{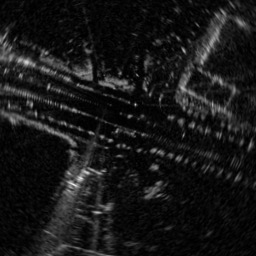}}\end{subfigure}
    \caption{}
    \end{subfigure}
    \begin{subfigure}{\figsizeL}
    \centering
    \begin{subfigure}{\figsize}{\includegraphics[width=\figsize]{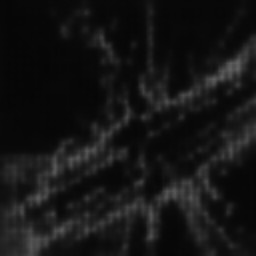}}\end{subfigure}
    \begin{subfigure}{\figsize}{\includegraphics[width=\figsize]{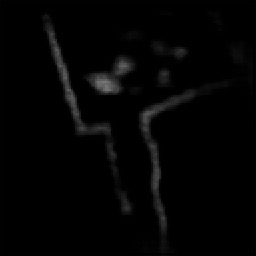}}\end{subfigure}
    \begin{subfigure}{\figsize}{\includegraphics[width=\figsize]{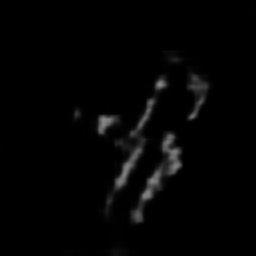}}\end{subfigure}
    \begin{subfigure}{\figsize}{\includegraphics[width=\figsize]{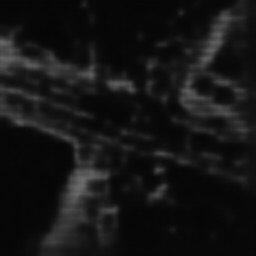}}\end{subfigure}
    \caption{}
    \end{subfigure}
    \begin{subfigure}{\figsizeL}
    \centering
    \begin{subfigure}{\figsize}{\includegraphics[width=\figsize]{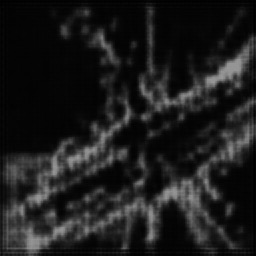}}\end{subfigure}
    \begin{subfigure}{\figsize}{\includegraphics[width=\figsize]{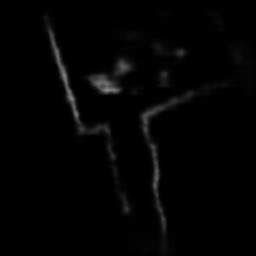}}\end{subfigure}
    \begin{subfigure}{\figsize}{\includegraphics[width=\figsize]{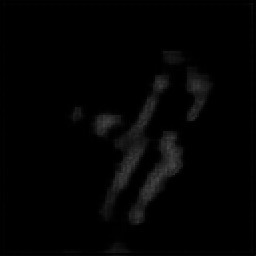}}\end{subfigure}
    \begin{subfigure}{\figsize}{\includegraphics[width=\figsize]{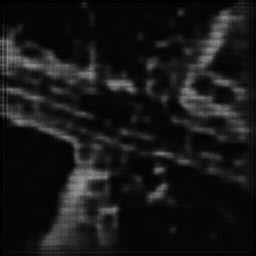}}\end{subfigure}
    \caption{}
    \end{subfigure}
    \caption{\label{fig:qualitative} \footnotesize Images at various stages of our method:
    map image $A$ (a), live data image $B$ (b), output of rotation inference $B_\theta$ (c), embedding $B^\dagger_\theta$ (d), pixel-wise aligned ground truth $B_{\theta, \alpha}$ (e), synthetic image $\tilde{B}_{\theta, \alpha}$ (f), embedding$\tilde{B}^\dagger_{\theta, \alpha}$ (g).
   From top to bottom: radar localisation against satellite imagery evaluated on MulRan, lidar localisation against satellite imagery evaluated on RobotCar and KITTI, radar localisation against lidar map evaluated on MulRan.
    \vspace{-4mm}
}

\end{figure*}

\begin{table}[!h]
\begin{tabular}{c|ccc|cc}
\hline
                & \multicolumn{3}{c|}{\textbf{Mean Error (metric)}} & \multicolumn{2}{c}{\textbf{(pixel)}} \\
                & $x~\mathrm{(m)}$               & $y~\mathrm{(m)}$              & $\theta~\mathrm{(\degree)}$               & $x$                      & $y$               \\ \hline
RobotCar (ours) &       2.21      &    2.57      &   2.65               &             2.55             &           2.97              \\
MulRan (ours)   &       3.57     &     3.26    &       2.15      &         4.53               &            4.13  \\   \hline
RobotCar (RSL-Net)  &       2.66      &    3.41      &   2.45 &             3.07              &          3.93             \\
MulRan (RSL-Net)  &         3.37     &     2.61     &       1.40      &         4.28              &               3.32	\\ \hline
 RobotCar (CycleGAN) &       6.41      &    9.05      &   2.65               &             7.40             &           10.44              \\
MulRan (CycleGAN)   &       4.84     &     4.39    &       2.15      &         6.14               &            5.58  \\  
\end{tabular}
\captionof{table}{\footnotesize \label{tab:results_rl} 
Mean error for radar localisation against prior lidar map.}
\end{table}

\label{sec:rll_experiment}
Though our method is designed for localising against satellite imagery, we show it can also handle more standard forms of cross-modality localisation.
Here we build a lidar map using a prior traversal, and localise using radar from a later traversal.

We demonstrate on the RobotCar and MulRan datasets, where we use the same resolution as in Section \ref{sec:rsl_experiment}.
For RobotCar, we use ground truth to build a lidar map from sequence no. 2.
Radar data in the training sections from no. 5 and no. 6 as in Figure \ref{fig:splits} form the training set, while the test section from sequence no. 5 forms the test set.
For MulRan, lidar maps are built from \texttt{KAIST 01} and \texttt{Sejong 01}, and we localised using radar data from \texttt{KAIST 02} and \texttt{Sejong 02}, which are split into training, validation, and test sets.
This resulted in a test set consisting of 201 frames from RobotCar and 272 frames from MulRan, spanning a total distance of near $4~\mathrm{km}.$
The localisation results are shown in Table \ref{tab:results_rl}.

This experiment is more suitable for naive image generation methods such as CycleGAN \cite{zhu2017unpaired} than previous experiments, as the field-of-view is considerably more compatible when both modalities are from range sensors.
In Table \ref{tab:results_rl}, we show results where we replaced the image generation stage of our method by CycleGAN, and kept other modules.
The localisation results are however much worse when modality $\mathcal{A}$ is satellite imagery, as shown qualitatively in Figure \ref{fig:cyclegan_qual}. 

\subsection{Online Pose-Tracking System}
In prior experiments we assumed place recognition is always available, providing a coarse initial estimate for every frame.
Here we present a stand-alone pose-tracking system by continuously localising against satellite imagery.
Given a coarse initial estimate (e.g., from GPS) for the first frame, the vehicle localises and computes its pose within the satellite map.
The initial estimate for every frame onward is then set to be the computed pose of the previous frame.
We only need place recognition once at the very beginning; the vehicle then tracks its pose onward without relying on any other measurements.

\subsubsection{Introspection}
As localising using satellite imagery is challenging, the result will not always be accurate.
Our method, however, naturally allows for introspection.
A synthetic image $\tilde{B}_{\theta, \alpha}$ was generated from $A$ and $B_\theta.$
We can apply a known small translation offset $\delta$ to $A$ to form $A_\delta.$
Taking $A_\delta$ and $B_\theta$ we can generate $\tilde{B}_{\theta, \alpha + \delta}.$
Finally, we can compute a translation offset $\hat{\delta}$ by passing $\tilde{B}_{\theta, \alpha + \delta}$ and $\tilde{B}_{\theta, \alpha}$ through the learned embeddings and maximising correlation.

Let $d_\mathrm{intro}  = \norm{\delta - \hat{\delta}}.$
A large value of $d_\mathrm{intro}$ indicates the generated images are erroneous.
This allows us to examine the solution quality; our system falls back to using odometry for dead-reckoning when $d_\mathrm{intro}$ exceeds a threshold.
We do not require high-quality odometry, but rather only use a naive approach by directly maximising correlation between two consecutive frames without any learned modules.
In our experiments, we set $\delta$ to be $\begin{bmatrix}
10 & 10
\end{bmatrix}^T,$ and $d_{\mathrm{intro}}$ to be 5.

\subsubsection{Results}
We conduct two experiments on the test set of RobotCar, one where we track a radar using satellite imagery, and one where we track a lidar.
For both experiments we run localisation at $4~\mathrm{Hz}.$
The results are shown in Figure \ref{fig:systems_result}.
If the solution error is too large, then the initial estimate will be too off for a sufficient overlap between the next queried satellite image and live data, resulting in losing track of the vehicle.
Though the solution error can be larger than $10~\mathrm{m}$ at times, our system  continuously localises the vehicle for over a kilometre without completely losing track.
For the lidar experiment, the solutions are sufficiently accurate to not require any odometry.
Our experiments are single-frame localisations, and we make no attempt at windowed/batch optimisation or loop closures.

\vspace{-1mm}
\subsection{Further Qualitative Results}
Additional qualitative results are presented in Figure \ref{fig:qualitative} showing various stages of our methods for different modalities.

\section{Conclusion and Future Work}
We present self-supervised learning to address cross-modality metric localisation between satellite imagery and on-board range sensors, without using metrically accurate ground truth for training.
Our method is validated across a large number of experiments for multiple modes of localisation, with results on-par with prior supervised approach.
A coarse initial pose estimate is needed for our method to compute metric localisation.
An extension would then be to solve place recognition for a range sensor within a large satellite map.

\section*{Acknowledgments}
We thank Giseop Kim from IRAP Lab, KAIST for providing GPS data for the MulRan Dataset.



\newpage
\bibliographystyle{plainnat}
\bibliography{main}

\newpage

\section*{Supplementary Material}
\subsection{Network Architecture}
Here we provide details on the network architecture of the various networks used in our method. 
We make use of the following abbreviations:
\begin{itemize}
    \item RP($p$): 2D reflection padding of $p$
    \item Conv($C_\mathrm{in}, C_\mathrm{out}, k, s, p$): convolution with $C_\mathrm{in}$ input channels, $C_\mathrm{out}$ output channels, kernel size $k,$ stride $s,$ padding $p,$ and bias
    \item IN: instance normalisation
    \item ReLU: rectified linear unit
    \item LReLU($ns$): leaky ReLU with negative slope ($ns$)
    \item Drop($d$): dropout with ratio $d$
    \item ConvT($C_\mathrm{in}, C_\mathrm{out}, k, s, p, p_\mathrm{out}$): transposed convolution with $C_\mathrm{in}$ input channels, $C_\mathrm{out}$ output channels, kernel size $k,$ stride $s,$ padding $p,$ output padding $ p_\mathrm{out},$ and bias
\end{itemize}

\begin{table}[!h]
\centering
\begin{tabular}{l}
\Xhline{3\arrayrulewidth}
Rotation Inference Function $f_R$ \\ 
Input shape - $n \times 4 \times 256 \times 256$ where $C = 4, H = W = 256$ \\ \hline
Conv(4, 32, 3, 2, 1) + IN + ReLU \\
Conv(32, 64, 3, 2, 1) + IN + ReLU \\
Conv(64, 128, 3, 2, 1) + IN + ReLU \\
Conv(128, 256, 3, 2, 1) + IN + ReLU \\
ResNet blocks ($\times 6$):\\
Conv(256, 256, 3, 1, 0) + IN + ReLU\\ 
Conv(256, 256, 3, 1, 0) + IN\\ \hline
Latent shape - $n \times 256  \times 16 \times 16$ \\ \hline
Sum along $C, H, W$ + Softmax + Reshape\\
Latent vector shape - $1 \times n,$ which are the softmax weights \\
Matrix-multiple softmax weights with the input\\ \hline
Shape of the multiplication product - $4 \times 256 \times 256$ \\
Extract the associated channel(s) to get $B_\theta$ (or $A_{\theta_i}$ during training)
\end{tabular}
\captionof{table}{\footnotesize \label{tab:rotation} Architecture for inferring rotation}
\end{table}

\begin{table}[!h]
\centering
\begin{tabular}{l}
\Xhline{3\arrayrulewidth}
Appearance Encoder $E_a$ \\ \hline
RP(3) + Conv(1, 16, 7, 1, 0) + IN + ReLU  \\
Conv(16, 32, 3, 2, 1) + IN + ReLU \\
Conv(32, 64, 3, 2, 1) + IN + ReLU \\
Conv(64, 128, 3, 2, 1) + IN + ReLU \\
Conv(128, 256, 3, 2, 1) + IN + ReLU \\ 
ResNet blocks ($\times 9$): \\ 
Conv(256, 256, 3, 1, 0) + IN + ReLU + Drop(0.5)\\ 
Conv(256, 256, 3, 1, 0) + IN\\ \hline

\Xhline{3\arrayrulewidth}
Intra-Modality Pose Encoder $E_p$ \\ \hline
RP(3) + Conv(2, 16, 7, 1, 0) + IN + ReLU  \\
Conv(16, 32, 3, 2, 1) + IN + ReLU \\
Conv(32, 64, 3, 2, 1) + IN + ReLU \\
Conv(64, 128, 3, 2, 1) + IN + ReLU \\
Conv(128, 256, 3, 2, 1) + IN + ReLU \\ 
ResNet blocks ($\times 9$): \\ 
Conv(256, 256, 3, 1, 0) + IN + ReLU + Drop(0.5)\\ 
Conv(256, 256, 3, 1, 0) + IN\\ \hline

\Xhline{3\arrayrulewidth}
Cross-Modality Pose Encoder $E_p^*$ \\ \hline
RP(3) + Conv(4, 16, 7, 1, 0) + IN + ReLU  \\
Conv(16, 32, 3, 2, 1) + IN + ReLU \\
Conv(32, 64, 3, 2, 1) + IN + ReLU \\
Conv(64, 128, 3, 2, 1) + IN + ReLU \\
Conv(128, 256, 3, 2, 1) + IN + ReLU \\ 
ResNet blocks ($\times 9$): \\ 
Conv(256, 256, 3, 1, 0) + IN + ReLU + Drop(0.5)\\ 
Conv(256, 256, 3, 1, 0) + IN\\ \hline

\Xhline{3\arrayrulewidth}
Decoder $D$ \\ \hline
ConvT(512, 256, 3, 2, 1, 1) + IN + ReLU + Drop(0.5)\\
ConvT(256, 128, 3, 2, 1, 1) + IN + ReLU + Drop(0.5)\\
ConvT(128, 64, 3, 2, 1, 1) + IN + ReLU + Drop(0.5)\\
ConvT(64, 32, 3, 2, 1, 1) + IN + ReLU + Drop(0.5)\\
RP(3) + Conv(32, 1, 7, 1, 0) + Sigmoid\\
\end{tabular}
\captionof{table}{\footnotesize \label{tab:generator} Architecture of for image generation}
\end{table}

\begin{table}[!h]
\centering
\begin{tabular}{l}
\Xhline{3\arrayrulewidth}
Encoder $E$ \\ \hline
RP(3) + Conv(1, 32, 7, 1, 0) + IN + ReLU  \\
Conv(32, 64, 3, 2, 1) + IN + ReLU \\
Conv(64, 128, 3, 2, 1) + IN + ReLU \\
Conv(128, 256, 3, 2, 1) + IN + ReLU \\
Conv(256, 512, 3, 2, 1) + IN + ReLU \\ 
ResNet blocks ($\times 9$): \\ 
Conv(512, 512, 3, 1, 0) + IN + ReLU + Drop(0.5)\\ 
Conv(512, 512, 3, 1, 0) + IN\\ \hline
\Xhline{3\arrayrulewidth}
Decoder $D$ \\ \hline
ConvT(512, 256, 3, 2, 1, 1) + IN + ReLU + Drop(0.5)\\
ConvT(256, 128, 3, 2, 1, 1) + IN + ReLU + Drop(0.5)\\
ConvT(128, 64, 3, 2, 1, 1) + IN + ReLU + Drop(0.5)\\
ConvT(64, 32, 3, 2, 1, 1) + IN + ReLU + Drop(0.5)\\
RP(3) + Conv(32, 1, 7, 1, 0) + Sigmoid\\
\end{tabular}
\captionof{table}{\footnotesize \label{tab:generator_s} Image generation for our implementation of RSL-Net [36]}
\end{table}

\begin{table}[!h]
\centering
\begin{tabular}{l}
\Xhline{3\arrayrulewidth}
Embedding Networks $H_B$ and $H_{\tilde{B}}$\\ \hline
Conv(1, 32, 4, 2, 0) \\
LReLU(0.2) + Conv(32, 64, 4, 2, 0) + IN \\
LReLU(0.2) + Conv(64, 128, 4, 2, 0) + IN\\
LReLU(0.2) + Conv(128, 256, 4, 2, 0) + IN\\
LReLU(0.2) + Conv(256, 512, 4, 2, 0) + IN\\
LReLU(0.2) + ReLU + Conv(512, 1024, 4, 2, 0) \\
ReLU + ConvT(1024, 512, 4, 2, 1, 0) + IN \\
ReLU + ConvT(512, 256, 4, 2, 1, 0) + IN \\
ReLU + ConvT(256, 128, 4, 2, 1, 0) + IN \\
ReLU + ConvT(128, 64, 4, 2, 1, 0) + IN \\
ReLU + ConvT(64, 32, 4, 2, 1, 0) + IN \\
ReLU + ConvT(32, 1, 4, 2, 1, 0) + Sigmoid \\
With skip connections in-between intermediate layers
\end{tabular}
\captionof{table}{\footnotesize \label{tab:embedding} U-Net architecture for learning embeddings}
\end{table}

The network architectures are shown in Tables \ref{tab:rotation} to \ref{tab:embedding}.
For comparison against prior supervised approach, we use the same architectures where possible.
We implemented the image generation network to have the same latent space size at the bottleneck, and the same number of down-samples and up-samples as in ours.

\subsection{Handling Larger Initial Offset}
Models for the experiments in Sections V-A, V-B, and V-C were trained assuming an initial translation offset in the range $[-25, 25]$ pixels, which corresponds to more than $[-20\mathrm{m}, 20\mathrm{m}]$ for the radar experiments.
In practice, the amount of offset our method can handle depends on the effective receptive field of the convolutional layers in the encoder and decoder networks for generating images.
If the offset is too large, the networks will not be able to encode and decode information needed to correctly generate $\tilde{B}_{\theta, \alpha}.$

\begin{figure*}[!t]
\vspace{2mm}
\centering
\begin{minipage}{.4\textwidth}
\centering
  \adjincludegraphics[height=7cm,clip]{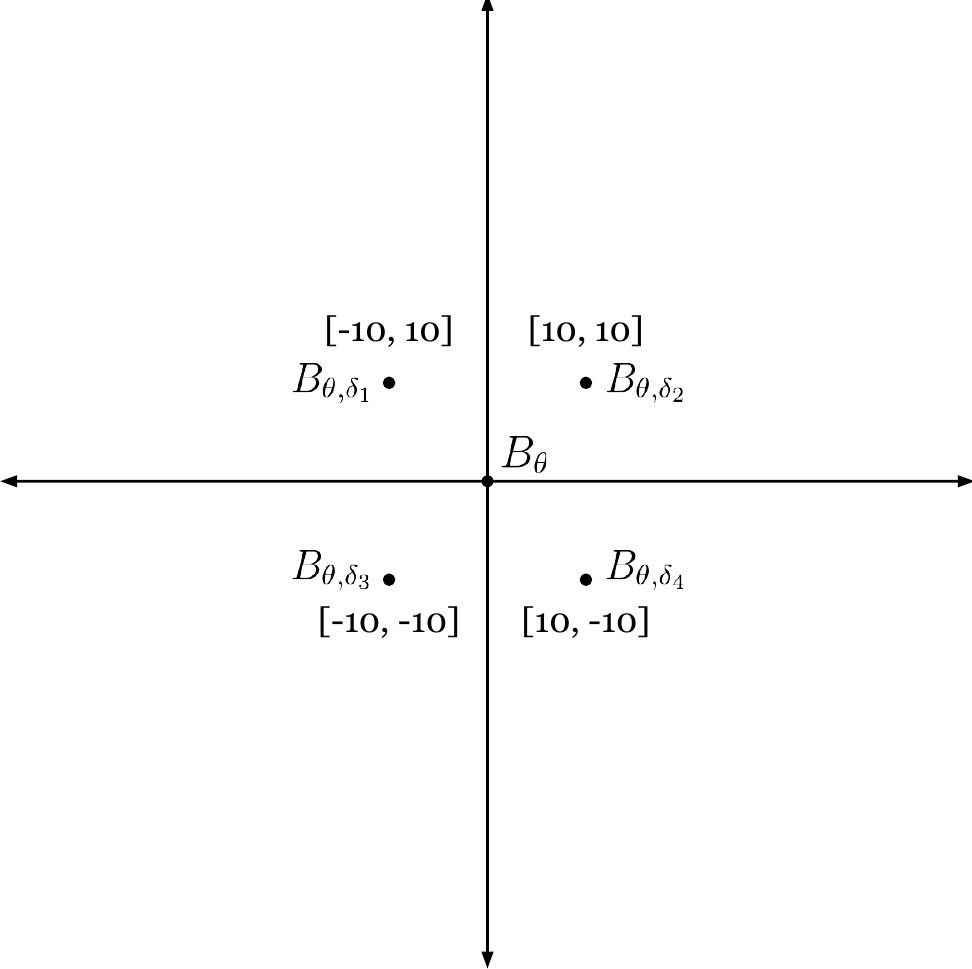}
  \captionsetup{justification=raggedright}
\caption{\label{fig:quadrant} \footnotesize The image $B_\theta$ is shifted into the four quadrants.
}
\end{minipage}%
\hspace{1cm}
\begin{minipage}{.4\textwidth}
\centering
    \adjincludegraphics[height=7cm,clip]{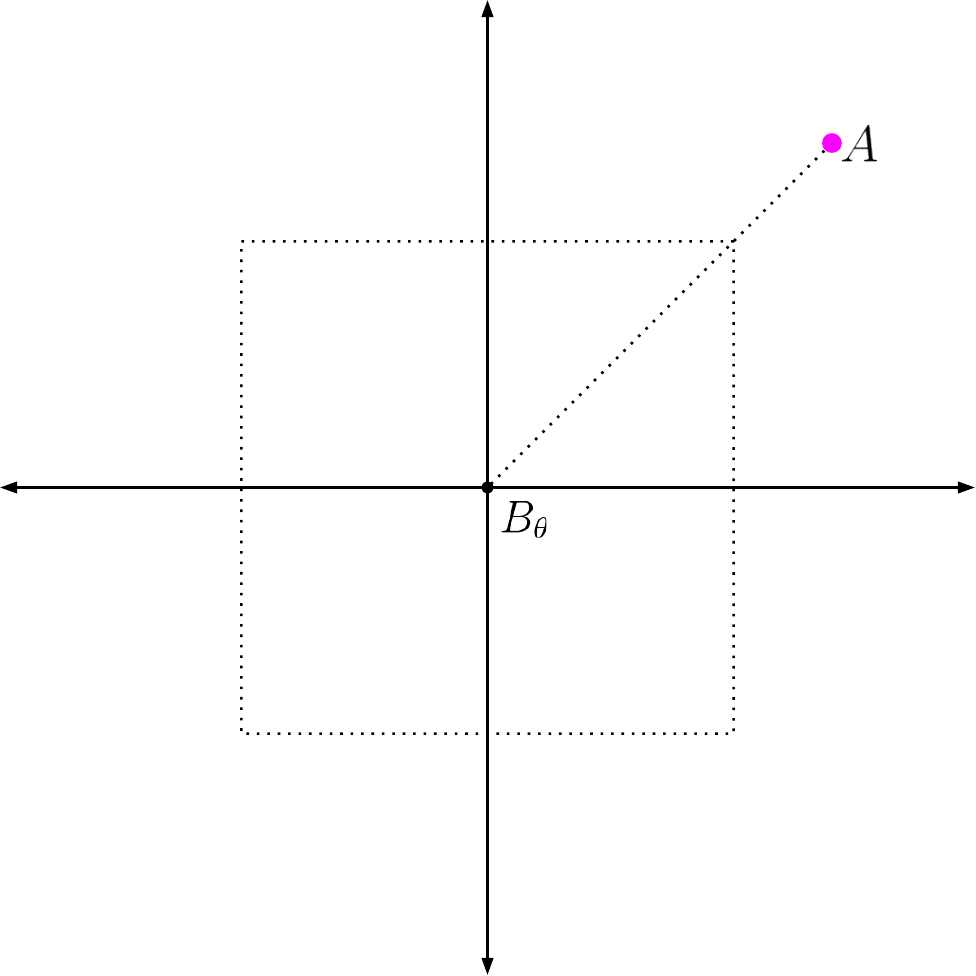}
    \captionof{figure}{\footnotesize\label{fig:large_offset_origin} The unknown translation offset $\alpha$ between $A$ and $B_\theta$ is larger than the networks are designed for.}
\end{minipage}
\end{figure*}

Our method, however, naturally allows for a strategy to deal with larger initial offsets, without needing to train different models.
At inference, rather than using just $B_\theta$ during image generation, we can apply known translation offsets $\delta_1, \delta_2, \delta_3,$ and $\delta_4$ to shift $B_\theta$ into each of the four quadrants.
This is depicted in Figure \ref{fig:quadrant}, where as an example, we shift $B_\theta$ by $[-10, 10], $ $[10, 10], $ $[-10, -10], $ and $[10, -10]$  pixels to form $B_{\theta, \delta_1}, $ $B_{\theta, \delta_2}, $ $B_{\theta, \delta_3},$ and $B_{\theta, \delta_4},$ respectively.

\begin{figure*}[!t]
\vspace{2mm}
\centering
\begin{minipage}{.4\textwidth}
\centering
  \adjincludegraphics[height=7cm,clip]{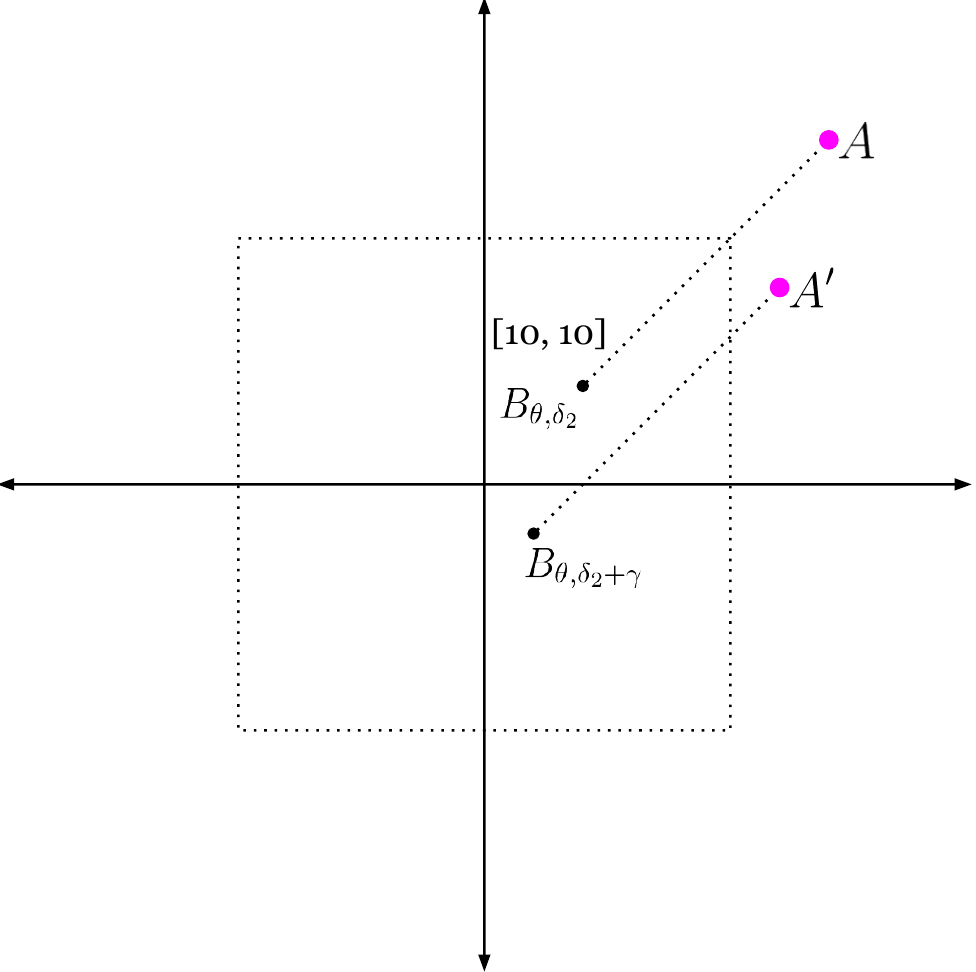}
  \captionsetup{justification=raggedright}
\caption{\label{fig:large_offset_2} \footnotesize If we shift $B_\theta$ by $[10, 10]$ to form $B_{\theta, \delta_2},$ then the offset between $A$ and $B_{\theta, \delta_2}$ is within what the networks are designed for.
In this case, generating $\tilde{B}_{\theta, \alpha}$ and $\tilde{B}_{\theta, \alpha + \gamma}$ should both be accurate, as the offset in both cases are within what the networks are trained for.
}
\end{minipage}%
\hspace{1cm}
\begin{minipage}{.4\textwidth}
\centering
    \adjincludegraphics[height=7cm,clip]{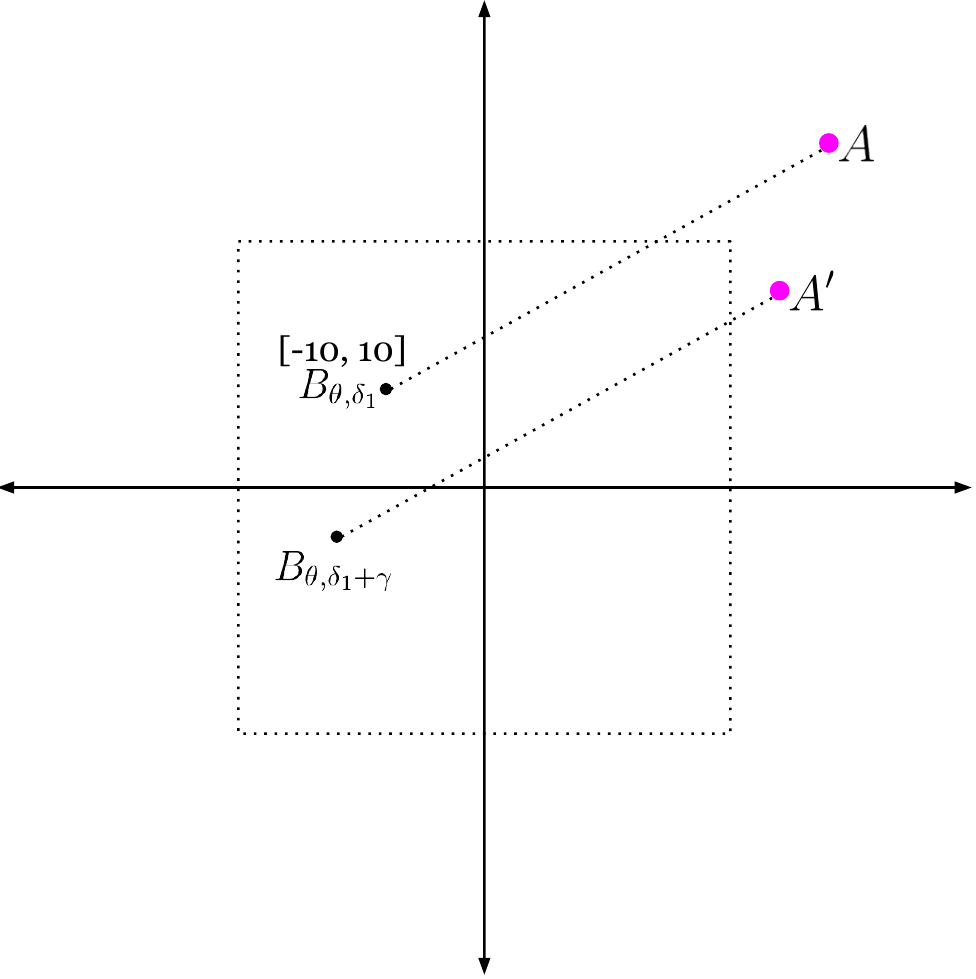}
    \captionof{figure}{\footnotesize\label{fig:large_offset_wrong} The resulting synthetic image will still be erroneous, if an incorrect quadrant is selected.
    Here the offset between $B_{\theta, \delta_1}$ and $A$ is larger than what the networks can handle.
    In this case, generating $\tilde{B}_{\theta, \alpha}$ and $\tilde{B}_{\theta, \alpha + \gamma}$ will both be problematic due to the issue with offsets.}
\end{minipage}
\end{figure*}

Figure \ref{fig:large_offset_origin} depicts a case where the translation $\alpha$ in the satellite image $A$ is $\alpha = \begin{bmatrix}
35 & 35
\end{bmatrix}^T$ pixels.
This is larger than the range our networks can handle, which is $[-25, 25],$ shown by the dashed box around the origin.
However, the offset between $B_{\theta, \delta_2}$ and $A$ is $\begin{bmatrix}
25 & 25
\end{bmatrix}^T,$ which is within  the range our networks can handle, as shown in Figure \ref{fig:large_offset_2}.

Forming $E_p^*(A, B_\theta)$ and $E_a(B_\theta)$ during image generation might lead to incorrect results when generating $\tilde{B}_{\theta, \alpha},$ as the offset between $A$ and $B_\theta$ is too large.
However, we can also generate $\tilde{B}_{\theta, \alpha}$ using $E_p^*(A, B_{\theta, \delta_2})$ and $E_a(B_{\theta, \delta_2}),$ and such combination does not suffer from the issue with large offsets.

The question is then which shifted image from $B_{\theta, \delta_1}, $ $B_{\theta, \delta_2}, $ $B_{\theta, \delta_3},$ and $B_{\theta, \delta_4}$ to choose from. 
Shown in Figure \ref{fig:large_offset_wrong}, generating $\tilde{B}_{\theta, \alpha}$ using $E_p^*(A, B_{\theta, \delta_1})$ and $E_a(B_{\theta, \delta_1})$ will also be problematic, as this combination suffers from the issue with large offsets.
The selection cannot be made ahead of the image generation as $\alpha$ is unknown.

We can generate five versions of $\tilde{B}_{\theta, \alpha}$ using $B_\theta,$ $B_{\theta, \delta_1}, $ $B_{\theta, \delta_2}, $ $B_{\theta, \delta_3},$ and $B_{\theta, \delta_4},$ and introspect the quality of each $\tilde{B}_{\theta, \alpha}.$
To do so, we apply a known shift $\gamma$ to $A$ to query for another image $A',$ and we can also shift each $B_{\theta, \delta_i}$ by $\gamma$ to form $B_{\theta, \delta_i + \gamma}$ (or $B_{\theta, \gamma}$ for $B_\theta$), as in Figures \ref{fig:large_offset_2} and \ref{fig:large_offset_wrong}.

For each shift $\delta_i$ (and zero shift for $B_\theta$), we can take the combination $E_p^*(A', B_{\theta, \delta_i + \gamma})$ and $E_a(B_{\theta, \delta_i + \gamma})$ to generate $\tilde{B}_{\theta, \alpha + \gamma},$ which should be pixel-wise aligned with $A'.$
If generating $\tilde{B}_{\theta, \alpha}$ is problematic due to large offsets, then so will generating $\tilde{B}_{\theta, \alpha + \gamma}$ be, as shown in Figure \ref{fig:large_offset_wrong}.
On the other hand, if the networks can correctly produce $\tilde{B}_{\theta, \alpha},$ they can also correctly produce $\tilde{B}_{\theta, \alpha + \gamma},$ as shown in Figure \ref{fig:large_offset_2}.

For each shift $\delta_i,$ we can compute a translation offset $\hat{\gamma}$ using $\tilde{B}_{\theta, \alpha + \gamma}$ and $\tilde{B}_{\theta, \alpha},$ along with an error term $e = \norm{\hat{\gamma} - \gamma}.$
For the five pairs of synthetic images, the one that results in the smallest $e$ will be used and passed downstream to solve for $\hat{\alpha}.$
This forms an augmented approach for handling initial offsets larger than what the models are trained for.

\begin{table}[!h]
\centering
\begin{tabular}{c|cc|cc}
\hline
                & \multicolumn{2}{c|}{\textbf{Mean Error (metric)}} & \multicolumn{2}{c}{\textbf{(pixel)}} \\
                & $x~\mathrm{(m)}$               & $y~\mathrm{(m)}$                         & $x$                      & $y$               \\ \hline
Direct &       6.62   &    7.88             &            7.64          &      9.09                 \\
Augmented &     4.67     &     5.54 &          5.39      &         6.40  \\  
Ours ($[-25, 25]$)&       3.44      &    5.40                  &             3.97              &           6.23              \\
\end{tabular}
\captionof{table}{\footnotesize \label{tab:large_offset} 
Radar localisation against satellite imagery evaluated on the test set of RobotCar, where the initial offset is in the range $[-35, 35]$ pixels.}
\end{table}

Table \ref{tab:large_offset} shows results on the RobotCar Dataset for radar localisation against satellite imagery, where the initial offset is now $[-35, 35]$ pixels.
Taking a model trained for an offset of $[-25, 25]$ and evaluate directly, the errors are high comparing to results in Sections V-A, V-B, and V-C.
However, taking our augmented approach by shifting $B_\theta$ and generating $\tilde{B}_{\theta, \alpha}$ multiple times, we can handle larger offsets without sacrificing significantly on accuracy.
The augmented method was not used in experiments shown in Section V due to the increased computational cost.

\subsection{Further Implementation Details}
Our method is implemented in PyTorch.
For training rotation inference $f_R$ and networks for image generation $E_a,$ $E_p,$ $E_p^*,$ and $D,$ we use a learning rate of $2\mathrm{e}^{-4}.$
For learning the embedding networks $H_B$ and $H_{\tilde{B}},$ we use a learning rate of $2\mathrm{e}^{-6}.$
We use Adam as the optimiser for all experiments.
The training is terminated when the validation loss increases for more than 5 epochs.
This results in approximately 80 to 150 epochs of training for learning  $f_R,$ $E_a,$ $E_p,$ $E_p^*,$ and $D,$ depending on the dataset and the specific experiment, and approximately 10 to 20 epochs for learning $H_B$ and $H_{\tilde{B}}.$

For the introspection method in Section V-D, we set $\delta$ to be $\begin{bmatrix}
10 & 10
\end{bmatrix}^T,$ and set the threshold for $d_{\mathrm{intro}}$ to be 5.
For rotation inference, we use an increment of $2\degree$ when forming the stack of rotated images.

\subsection{Ablation Study}
We perform ablation study to investigate the effect of reduced training data.
For radar localisation against satellite imagery on the RobotCar Dataset, we trained a model using approximately the first $20\%$ of training data, and another using every $10^{\mathrm{th}}$ frame of training data.
The results are shown in Table \ref{tab:ablation}.

\begin{table}[!h]
\centering
\begin{tabular}{c|ccc|cc}
\hline
                & \multicolumn{3}{c|}{\textbf{Mean Error (metric)}} & \multicolumn{2}{c}{\textbf{(pixel)}} \\
                & $x~\mathrm{(m)}$               & $y~\mathrm{(m)}$              & $\theta~\mathrm{(\degree)}$               & $x$                      & $y$               \\ \hline
RobotCar (full) &       3.44      &    5.40      &   3.03               &             3.97              &           6.23       \\ 
RobotCar (first $20\%$ ) &     7.96      &    7.45      &   6.03               &             9.18              &          8.59              \\ 
 RobotCar (every $10 ^\mathrm{th}$ ) &     4.36      &    6.18    &  4.40               &             5.03              &          7.14              \\ 
\end{tabular}
\captionof{table}{\footnotesize \label{tab:ablation} 
Ablation study for using reduced training data, evaluated on radar localisation against satellite imagery on the RobotCar Dataset.}
\end{table}

By using every $10 ^\mathrm{th}$ we have used only $10 \%$ of training data.
However, by sampling the data uniformly, we have a training set that is more varied
than selecting the first $20 \%,$ and therefore led to better performance.

\subsection{Additional Qualitative Results}
\begin{figure*}[h]%
    \centering
    \begin{subfigure}{\figsizeL}
    \centering
    \begin{subfigure}{\figsize}{\includegraphics[width=\figsize]{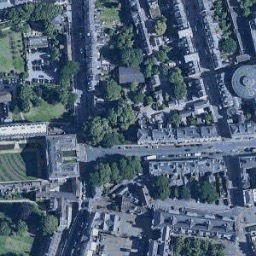}}\end{subfigure}
    \begin{subfigure}{\figsize}{\includegraphics[width=\figsize]{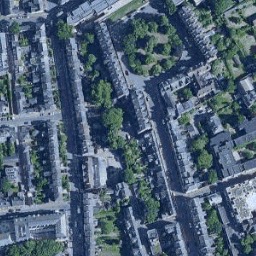}}\end{subfigure}
    \begin{subfigure}{\figsize}{\includegraphics[width=\figsize]{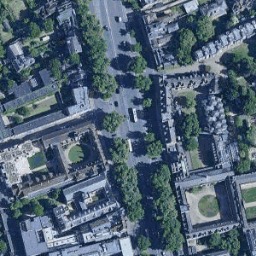}}\end{subfigure}
    \begin{subfigure}{\figsize}{\includegraphics[width=\figsize]{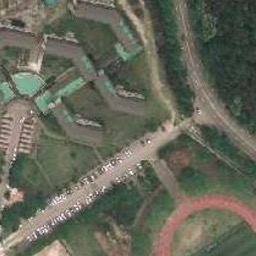}}\end{subfigure}
    \begin{subfigure}{\figsize}{\includegraphics[width=\figsize]{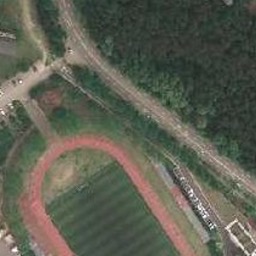}}\end{subfigure}
    \begin{subfigure}{\figsize}{\includegraphics[width=\figsize]{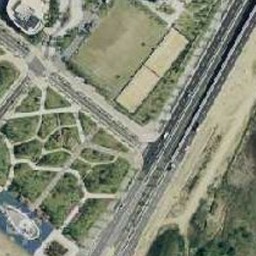}}\end{subfigure}
    \begin{subfigure}{\figsize}{\includegraphics[width=\figsize]{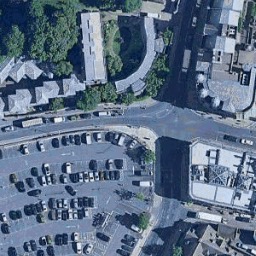}}\end{subfigure}
    \begin{subfigure}{\figsize}{\includegraphics[width=\figsize]{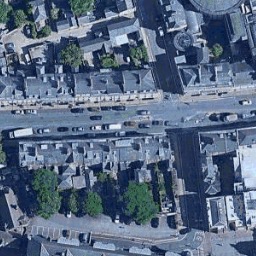}}\end{subfigure}
    \begin{subfigure}{\figsize}{\includegraphics[width=\figsize]{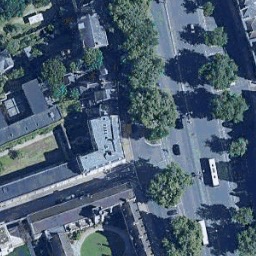}}\end{subfigure}
    \caption{}
    \end{subfigure}
    \begin{subfigure}{\figsizeL}
    \centering
    \begin{subfigure}{\figsize}{\includegraphics[width=\figsize]{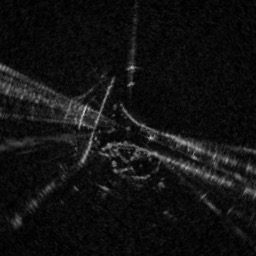}}\end{subfigure}
    \begin{subfigure}{\figsize}{\includegraphics[width=\figsize]{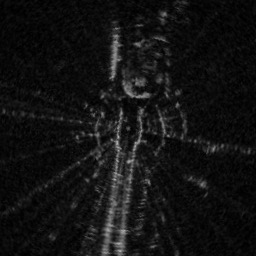}}\end{subfigure}
    \begin{subfigure}{\figsize}{\includegraphics[width=\figsize]{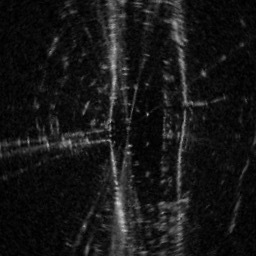}}\end{subfigure}
    \begin{subfigure}{\figsize}{\includegraphics[width=\figsize]{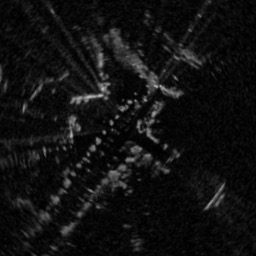}}\end{subfigure}
    \begin{subfigure}{\figsize}{\includegraphics[width=\figsize]{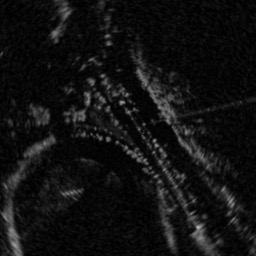}}\end{subfigure}
    \begin{subfigure}{\figsize}{\includegraphics[width=\figsize]{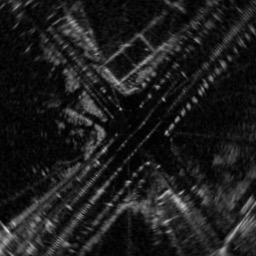}}\end{subfigure}
    \begin{subfigure}{\figsize}{\includegraphics[width=\figsize]{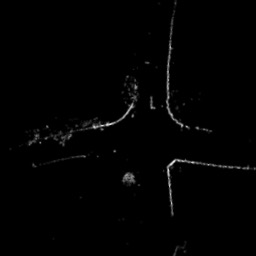}}\end{subfigure}
    \begin{subfigure}{\figsize}{\includegraphics[width=\figsize]{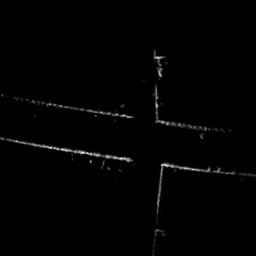}}\end{subfigure}
    \begin{subfigure}{\figsize}{\includegraphics[width=\figsize]{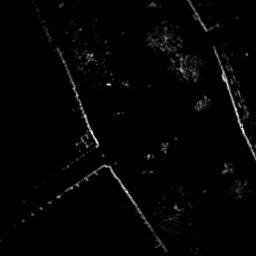}}\end{subfigure}
    \caption{}
    \end{subfigure}
    \begin{subfigure}{\figsizeL}
    \centering
    \begin{subfigure}{\figsize}{\includegraphics[width=\figsize]{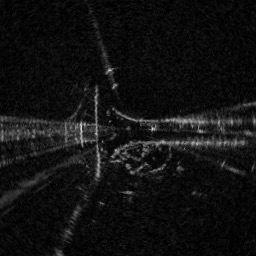}}\end{subfigure}
    \begin{subfigure}{\figsize}{\includegraphics[width=\figsize]{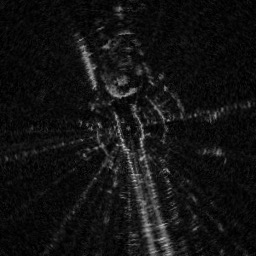}}\end{subfigure}
    \begin{subfigure}{\figsize}{\includegraphics[width=\figsize]{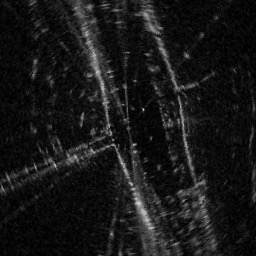}}\end{subfigure}
    \begin{subfigure}{\figsize}{\includegraphics[width=\figsize]{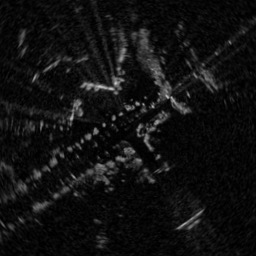}}\end{subfigure}
    \begin{subfigure}{\figsize}{\includegraphics[width=\figsize]{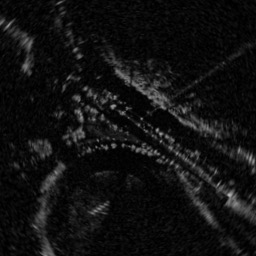}}\end{subfigure}
    \begin{subfigure}{\figsize}{\includegraphics[width=\figsize]{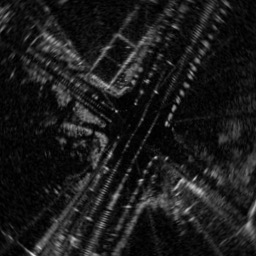}}\end{subfigure}
    \begin{subfigure}{\figsize}{\includegraphics[width=\figsize]{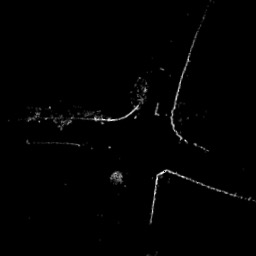}}\end{subfigure}
    \begin{subfigure}{\figsize}{\includegraphics[width=\figsize]{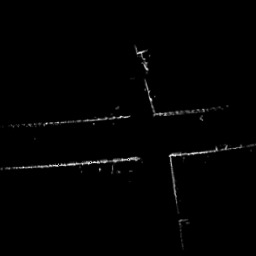}}\end{subfigure}
     \begin{subfigure}{\figsize}{\includegraphics[width=\figsize]{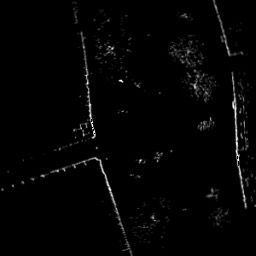}}\end{subfigure}
    \caption{}
    \end{subfigure}
    \begin{subfigure}{\figsizeL}
    \centering
    \begin{subfigure}{\figsize}{\includegraphics[width=\figsize]{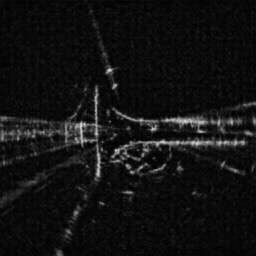}}\end{subfigure}
    \begin{subfigure}{\figsize}{\includegraphics[width=\figsize]{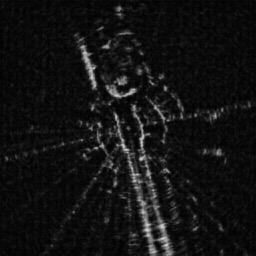}}\end{subfigure}
    \begin{subfigure}{\figsize}{\includegraphics[width=\figsize]{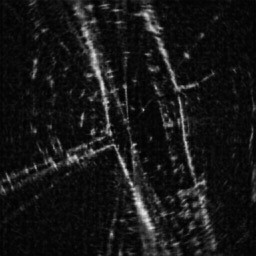}}\end{subfigure}
    \begin{subfigure}{\figsize}{\includegraphics[width=\figsize]{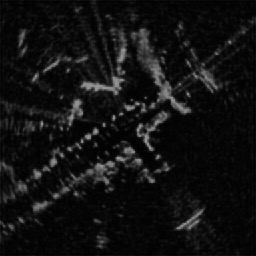}}\end{subfigure}
    \begin{subfigure}{\figsize}{\includegraphics[width=\figsize]{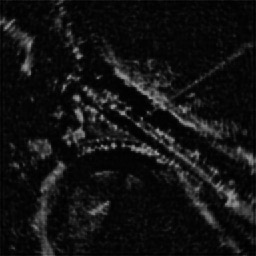}}\end{subfigure}
    \begin{subfigure}{\figsize}{\includegraphics[width=\figsize]{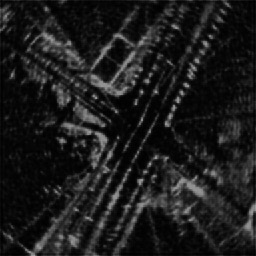}}\end{subfigure}
    \begin{subfigure}{\figsize}{\includegraphics[width=\figsize]{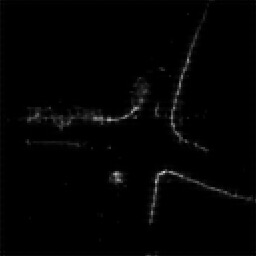}}\end{subfigure}
    \begin{subfigure}{\figsize}{\includegraphics[width=\figsize]{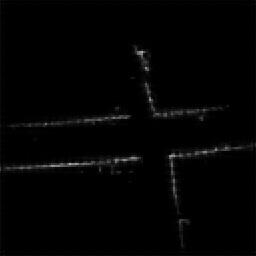}}\end{subfigure}
    \begin{subfigure}{\figsize}{\includegraphics[width=\figsize]{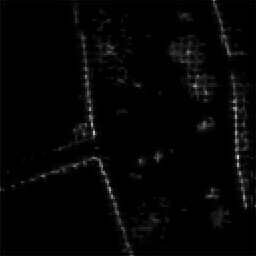}}\end{subfigure}
    \caption{}
    \end{subfigure}
    \begin{subfigure}{\figsizeL}
    \centering
    \begin{subfigure}{\figsize}{\includegraphics[width=\figsize]{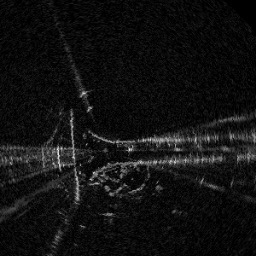}}\end{subfigure}
    \begin{subfigure}{\figsize}{\includegraphics[width=\figsize]{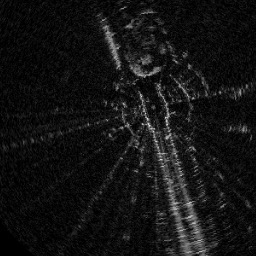}}\end{subfigure}
    \begin{subfigure}{\figsize}{\includegraphics[width=\figsize]{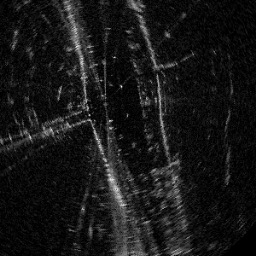}}\end{subfigure}
    \begin{subfigure}{\figsize}{\includegraphics[width=\figsize]{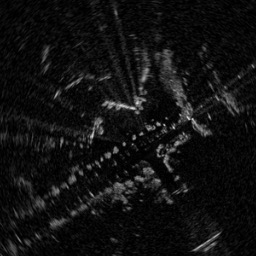}}\end{subfigure}
    \begin{subfigure}{\figsize}{\includegraphics[width=\figsize]{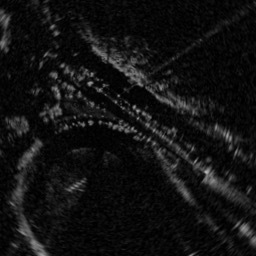}}\end{subfigure}
    \begin{subfigure}{\figsize}{\includegraphics[width=\figsize]{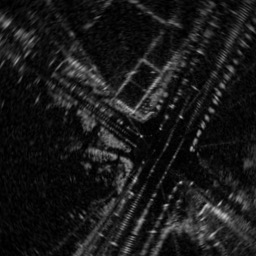}}\end{subfigure}
    \begin{subfigure}{\figsize}{\includegraphics[width=\figsize]{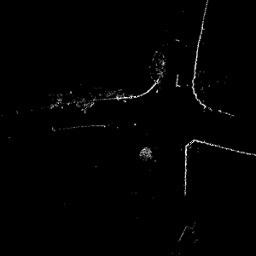}}\end{subfigure}
    \begin{subfigure}{\figsize}{\includegraphics[width=\figsize]{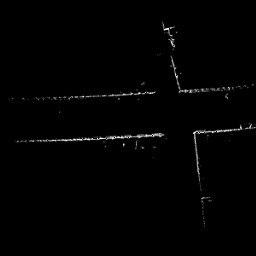}}\end{subfigure}
    \begin{subfigure}{\figsize}{\includegraphics[width=\figsize]{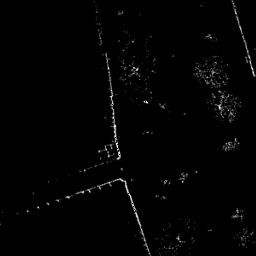}}\end{subfigure}
    \caption{}
    \end{subfigure}
    \begin{subfigure}{\figsizeL}
    \centering
    \begin{subfigure}{\figsize}{\includegraphics[width=\figsize]{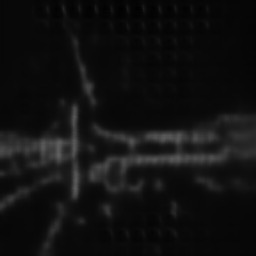}}\end{subfigure}
    \begin{subfigure}{\figsize}{\includegraphics[width=\figsize]{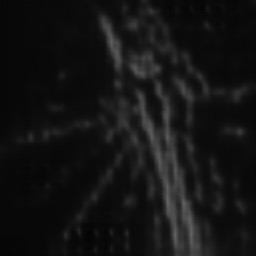}}\end{subfigure}
    \begin{subfigure}{\figsize}{\includegraphics[width=\figsize]{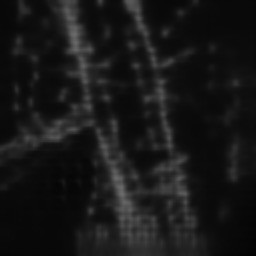}}\end{subfigure}
    \begin{subfigure}{\figsize}{\includegraphics[width=\figsize]{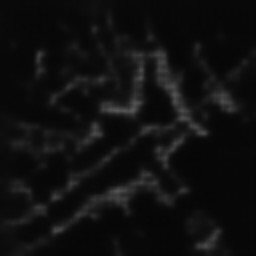}}\end{subfigure}
    \begin{subfigure}{\figsize}{\includegraphics[width=\figsize]{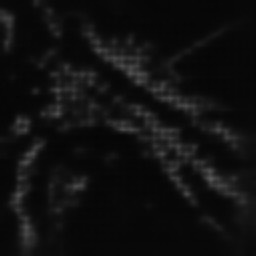}}\end{subfigure}
    \begin{subfigure}{\figsize}{\includegraphics[width=\figsize]{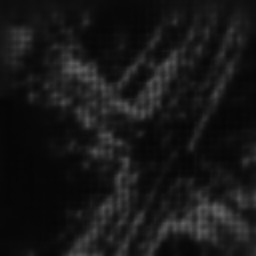}}\end{subfigure}
    \begin{subfigure}{\figsize}{\includegraphics[width=\figsize]{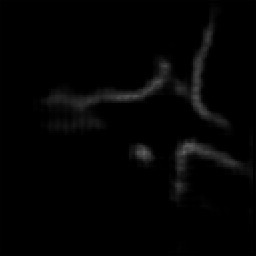}}\end{subfigure}
    \begin{subfigure}{\figsize}{\includegraphics[width=\figsize]{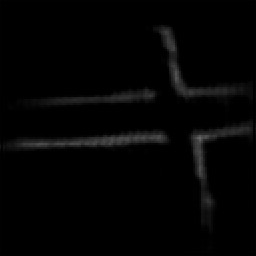}}\end{subfigure}
    \begin{subfigure}{\figsize}{\includegraphics[width=\figsize]{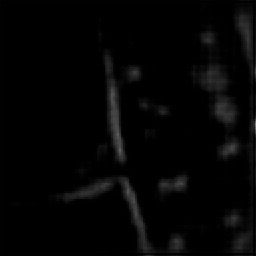}}\end{subfigure}
    \caption{}
    \end{subfigure}
    \begin{subfigure}{\figsizeL}
    \centering
    \begin{subfigure}{\figsize}{\includegraphics[width=\figsize]{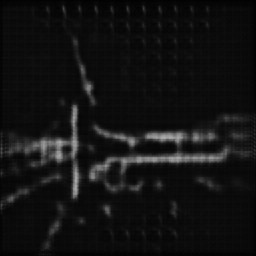}}\end{subfigure}
    \begin{subfigure}{\figsize}{\includegraphics[width=\figsize]{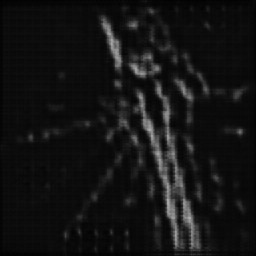}}\end{subfigure}
    \begin{subfigure}{\figsize}{\includegraphics[width=\figsize]{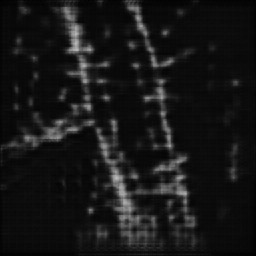}}\end{subfigure}
    \begin{subfigure}{\figsize}{\includegraphics[width=\figsize]{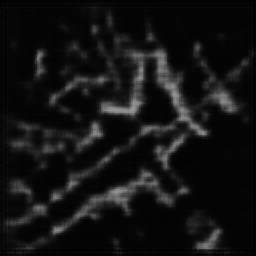}}\end{subfigure}
    \begin{subfigure}{\figsize}{\includegraphics[width=\figsize]{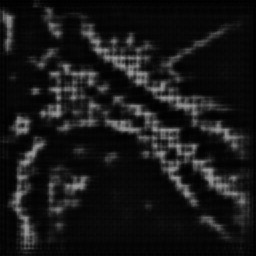}}\end{subfigure}
    \begin{subfigure}{\figsize}{\includegraphics[width=\figsize]{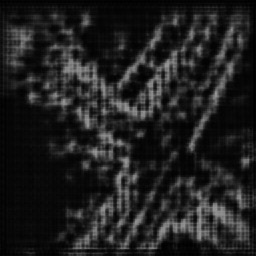}}\end{subfigure}
    \begin{subfigure}{\figsize}{\includegraphics[width=\figsize]{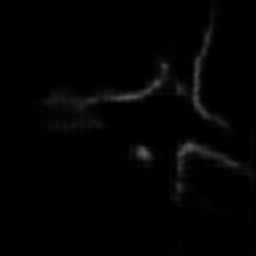}}\end{subfigure}
    \begin{subfigure}{\figsize}{\includegraphics[width=\figsize]{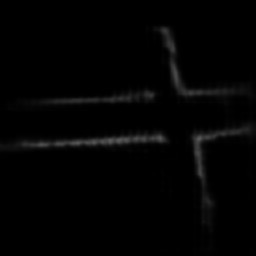}}\end{subfigure}
     \begin{subfigure}{\figsize}{\includegraphics[width=\figsize]{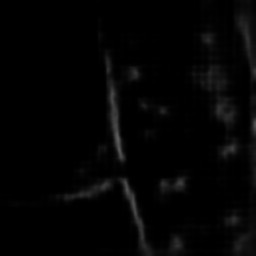}}\end{subfigure}
    \caption{}
    \end{subfigure}
    \caption{\label{fig:qualitative_sm1} \footnotesize Images at various stages of our method:
    map image $A$ (a), live data image $B$ (b), output of rotation inference $B_\theta$ (c), embedding $B^\dagger_\theta$ (d), pixel-wise aligned ground truth $B_{\theta, \alpha}$ (e), synthetic image $\tilde{B}_{\theta, \alpha}$ (f), embedding$\tilde{B}^\dagger_{\theta, \alpha}$ (g).
    From top to bottom: radar localisation against satellite imagery evaluated on RobotCar (rows 1-3) and MulRan (rows 4-6), lidar localisation against satellite imagery evaluated on RobotCar (rows 7-9).
    \vspace{-2mm}
}
\end{figure*}

\begin{figure*}[h]%
    \centering
    \begin{subfigure}{\figsizeL}
    \centering
    \begin{subfigure}{\figsize}{\includegraphics[width=\figsize]{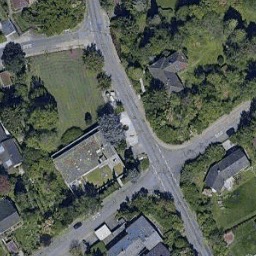}}\end{subfigure}
    \begin{subfigure}{\figsize}{\includegraphics[width=\figsize]{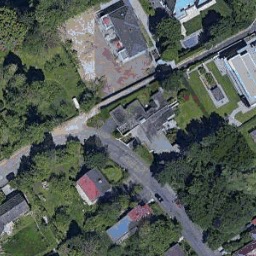}}\end{subfigure}
    \begin{subfigure}{\figsize}{\includegraphics[width=\figsize]{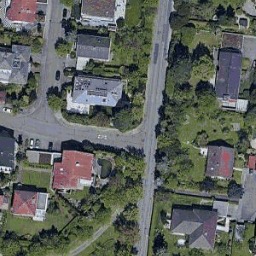}}\end{subfigure}
    \begin{subfigure}{\figsize}{\includegraphics[width=\figsize]{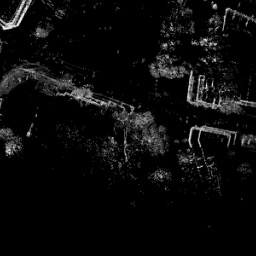}}\end{subfigure}
    \begin{subfigure}{\figsize}{\includegraphics[width=\figsize]{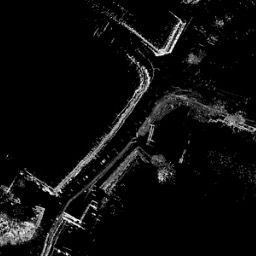}}\end{subfigure}
    \begin{subfigure}{\figsize}{\includegraphics[width=\figsize]{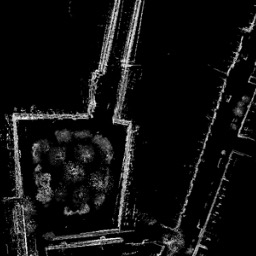}}\end{subfigure}
    \begin{subfigure}{\figsize}{\includegraphics[width=\figsize]{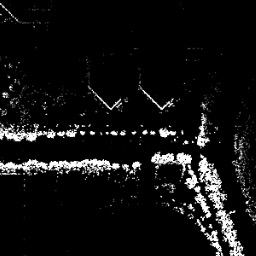}}\end{subfigure}
    \begin{subfigure}{\figsize}{\includegraphics[width=\figsize]{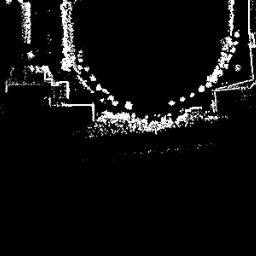}}\end{subfigure}
    \begin{subfigure}{\figsize}{\includegraphics[width=\figsize]{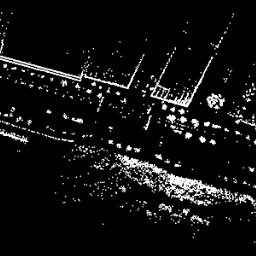}}\end{subfigure}
    \caption{}
    \end{subfigure}
    \begin{subfigure}{\figsizeL}
    \centering
    \begin{subfigure}{\figsize}{\includegraphics[width=\figsize]{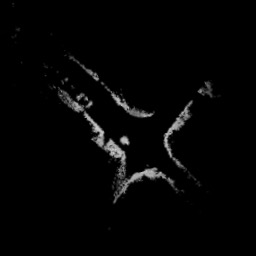}}\end{subfigure}
    \begin{subfigure}{\figsize}{\includegraphics[width=\figsize]{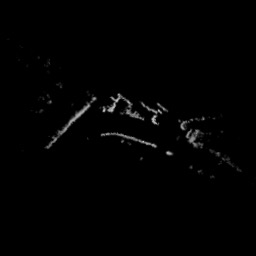}}\end{subfigure}
    \begin{subfigure}{\figsize}{\includegraphics[width=\figsize]{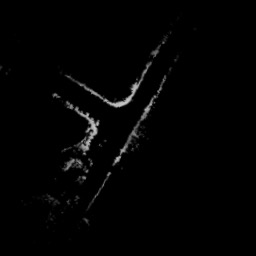}}\end{subfigure}
    \begin{subfigure}{\figsize}{\includegraphics[width=\figsize]{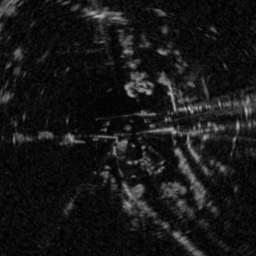}}\end{subfigure}
    \begin{subfigure}{\figsize}{\includegraphics[width=\figsize]{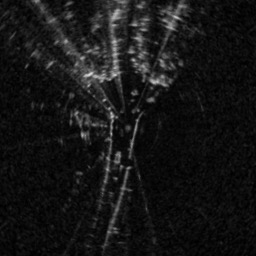}}\end{subfigure}
    \begin{subfigure}{\figsize}{\includegraphics[width=\figsize]{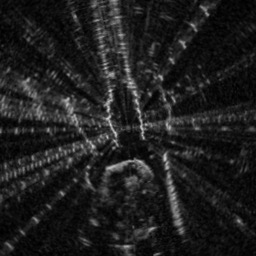}}\end{subfigure}
    \begin{subfigure}{\figsize}{\includegraphics[width=\figsize]{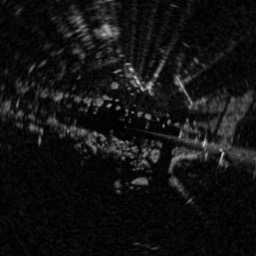}}\end{subfigure}
    \begin{subfigure}{\figsize}{\includegraphics[width=\figsize]{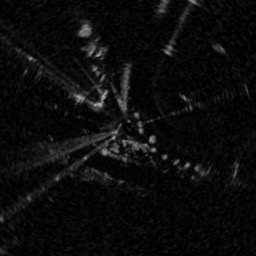}}\end{subfigure}
    \begin{subfigure}{\figsize}{\includegraphics[width=\figsize]{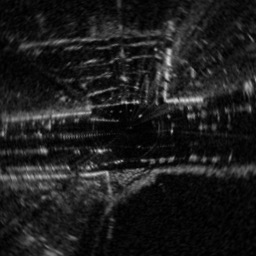}}\end{subfigure}
    \caption{}
    \end{subfigure}
    \begin{subfigure}{\figsizeL}
    \centering
    \begin{subfigure}{\figsize}{\includegraphics[width=\figsize]{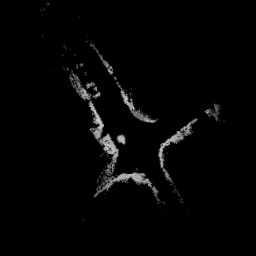}}\end{subfigure}
    \begin{subfigure}{\figsize}{\includegraphics[width=\figsize]{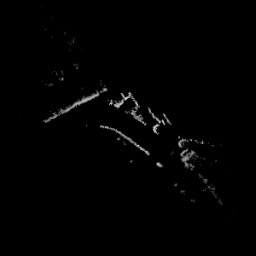}}\end{subfigure}
    \begin{subfigure}{\figsize}{\includegraphics[width=\figsize]{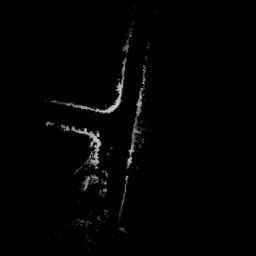}}\end{subfigure}
    \begin{subfigure}{\figsize}{\includegraphics[width=\figsize]{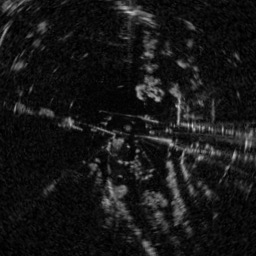}}\end{subfigure}
    \begin{subfigure}{\figsize}{\includegraphics[width=\figsize]{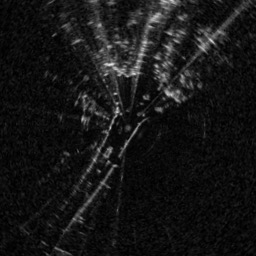}}\end{subfigure}
    \begin{subfigure}{\figsize}{\includegraphics[width=\figsize]{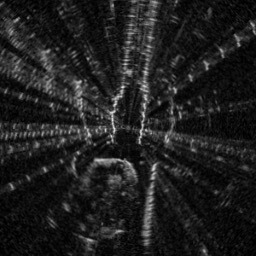}}\end{subfigure}
    \begin{subfigure}{\figsize}{\includegraphics[width=\figsize]{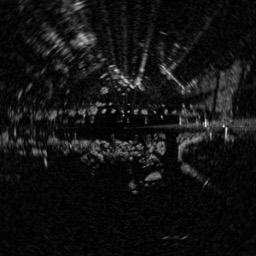}}\end{subfigure}
    \begin{subfigure}{\figsize}{\includegraphics[width=\figsize]{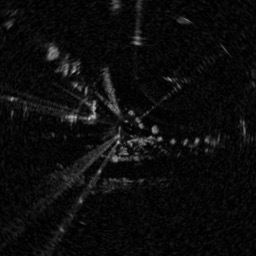}}\end{subfigure}
     \begin{subfigure}{\figsize}{\includegraphics[width=\figsize]{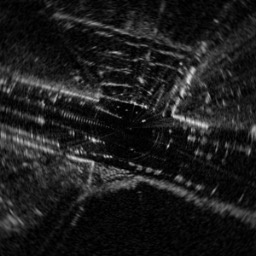}}\end{subfigure}
    \caption{}
    \end{subfigure}
    \begin{subfigure}{\figsizeL}
    \centering
    \begin{subfigure}{\figsize}{\includegraphics[width=\figsize]{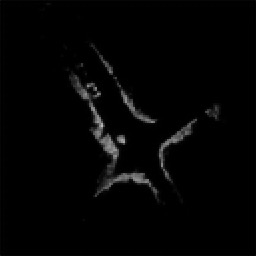}}\end{subfigure}
    \begin{subfigure}{\figsize}{\includegraphics[width=\figsize]{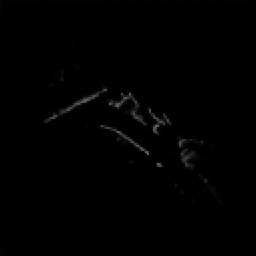}}\end{subfigure}
    \begin{subfigure}{\figsize}{\includegraphics[width=\figsize]{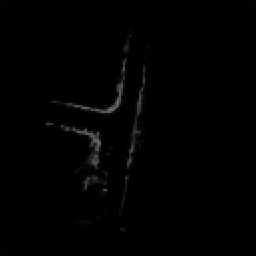}}\end{subfigure}
    \begin{subfigure}{\figsize}{\includegraphics[width=\figsize]{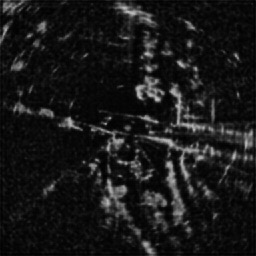}}\end{subfigure}
    \begin{subfigure}{\figsize}{\includegraphics[width=\figsize]{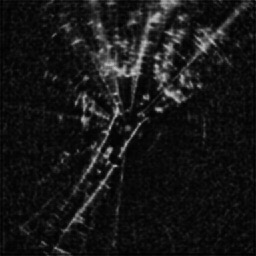}}\end{subfigure}
    \begin{subfigure}{\figsize}{\includegraphics[width=\figsize]{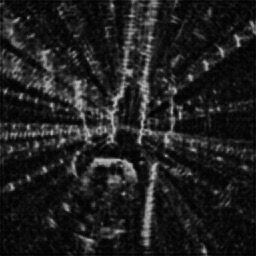}}\end{subfigure}
    \begin{subfigure}{\figsize}{\includegraphics[width=\figsize]{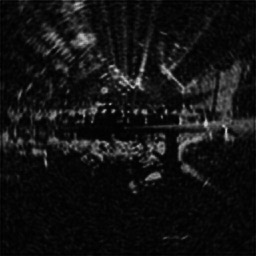}}\end{subfigure}
    \begin{subfigure}{\figsize}{\includegraphics[width=\figsize]{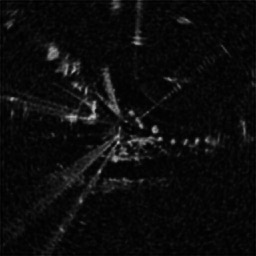}}\end{subfigure}
    \begin{subfigure}{\figsize}{\includegraphics[width=\figsize]{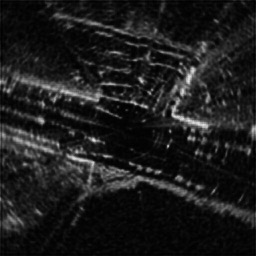}}\end{subfigure}
    \caption{}
    \end{subfigure}
    \begin{subfigure}{\figsizeL}
    \centering
    \begin{subfigure}{\figsize}{\includegraphics[width=\figsize]{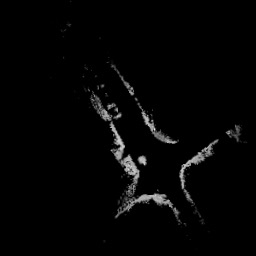}}\end{subfigure}
    \begin{subfigure}{\figsize}{\includegraphics[width=\figsize]{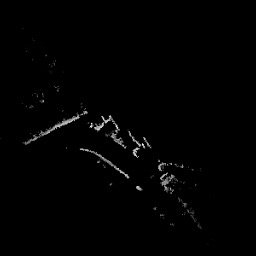}}\end{subfigure}
    \begin{subfigure}{\figsize}{\includegraphics[width=\figsize]{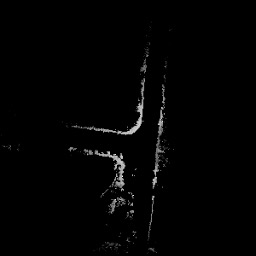}}\end{subfigure}
    \begin{subfigure}{\figsize}{\includegraphics[width=\figsize]{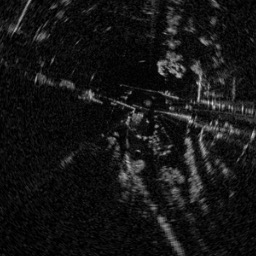}}\end{subfigure}
    \begin{subfigure}{\figsize}{\includegraphics[width=\figsize]{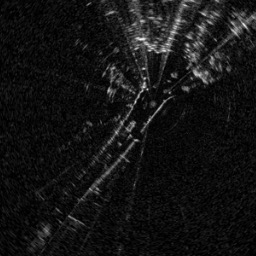}}\end{subfigure}
    \begin{subfigure}{\figsize}{\includegraphics[width=\figsize]{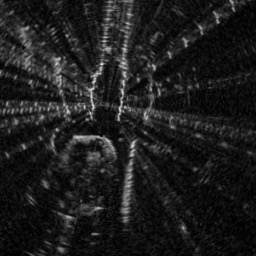}}\end{subfigure}
    \begin{subfigure}{\figsize}{\includegraphics[width=\figsize]{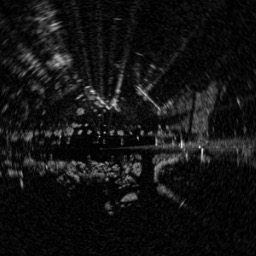}}\end{subfigure}
    \begin{subfigure}{\figsize}{\includegraphics[width=\figsize]{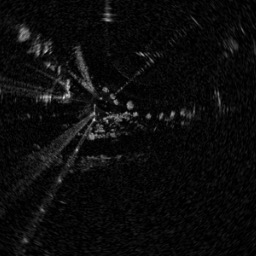}}\end{subfigure}
    \begin{subfigure}{\figsize}{\includegraphics[width=\figsize]{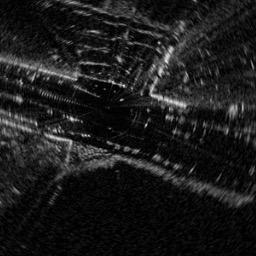}}\end{subfigure}
    \caption{}
    \end{subfigure}
    \begin{subfigure}{\figsizeL}
    \centering
    \begin{subfigure}{\figsize}{\includegraphics[width=\figsize]{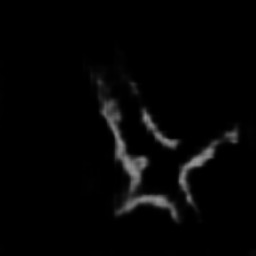}}\end{subfigure}
    \begin{subfigure}{\figsize}{\includegraphics[width=\figsize]{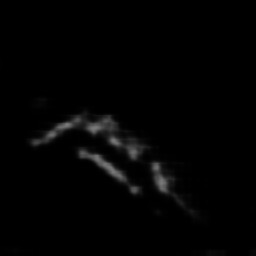}}\end{subfigure}
    \begin{subfigure}{\figsize}{\includegraphics[width=\figsize]{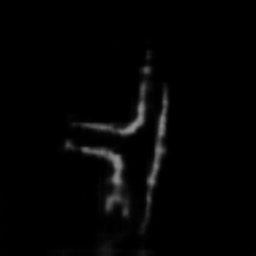}}\end{subfigure}
    \begin{subfigure}{\figsize}{\includegraphics[width=\figsize]{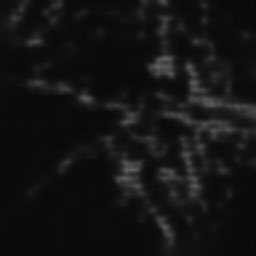}}\end{subfigure}
    \begin{subfigure}{\figsize}{\includegraphics[width=\figsize]{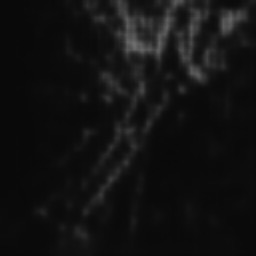}}\end{subfigure}
    \begin{subfigure}{\figsize}{\includegraphics[width=\figsize]{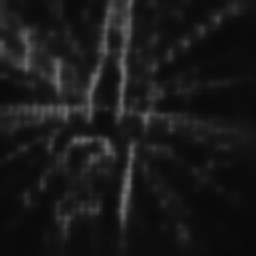}}\end{subfigure}
    \begin{subfigure}{\figsize}{\includegraphics[width=\figsize]{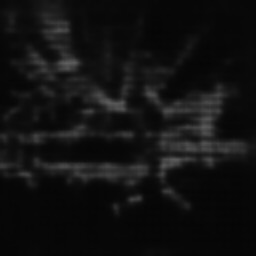}}\end{subfigure}
    \begin{subfigure}{\figsize}{\includegraphics[width=\figsize]{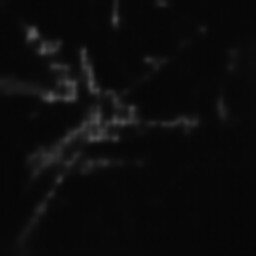}}\end{subfigure}
    \begin{subfigure}{\figsize}{\includegraphics[width=\figsize]{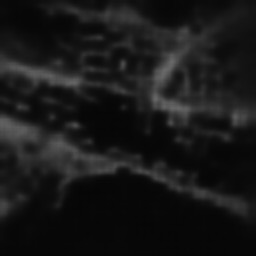}}\end{subfigure}
    \caption{}
    \end{subfigure}
    \begin{subfigure}{\figsizeL}
    \centering
    \begin{subfigure}{\figsize}{\includegraphics[width=\figsize]{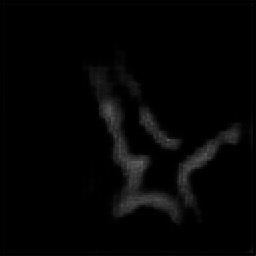}}\end{subfigure}
    \begin{subfigure}{\figsize}{\includegraphics[width=\figsize]{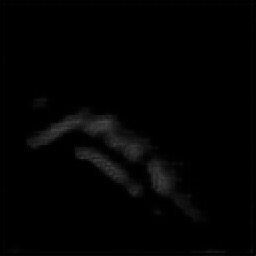}}\end{subfigure}
    \begin{subfigure}{\figsize}{\includegraphics[width=\figsize]{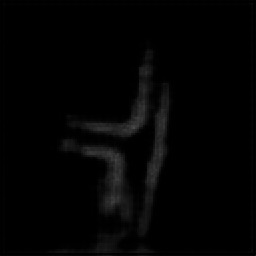}}\end{subfigure}
    \begin{subfigure}{\figsize}{\includegraphics[width=\figsize]{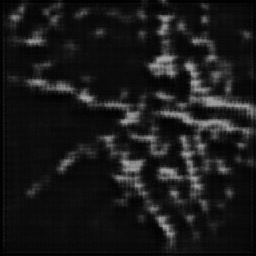}}\end{subfigure}
    \begin{subfigure}{\figsize}{\includegraphics[width=\figsize]{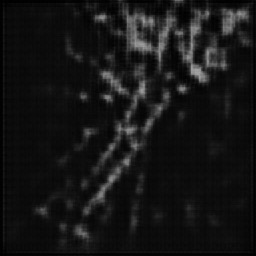}}\end{subfigure}
    \begin{subfigure}{\figsize}{\includegraphics[width=\figsize]{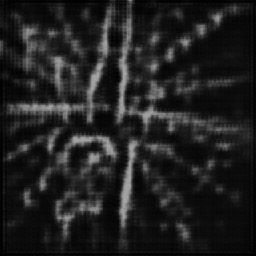}}\end{subfigure}
    \begin{subfigure}{\figsize}{\includegraphics[width=\figsize]{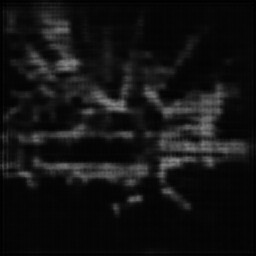}}\end{subfigure}
    \begin{subfigure}{\figsize}{\includegraphics[width=\figsize]{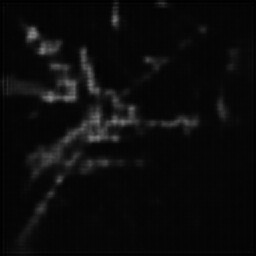}}\end{subfigure}
     \begin{subfigure}{\figsize}{\includegraphics[width=\figsize]{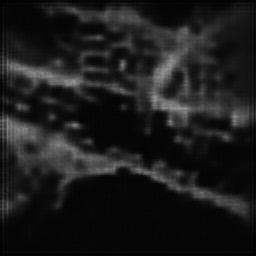}}\end{subfigure}
    \caption{}
    \end{subfigure}
    \caption{\label{fig:qualitative_sm2} \footnotesize Images at various stages of our method:
    map image $A$ (a), live data image $B$ (b), output of rotation inference $B_\theta$ (c), embedding $B^\dagger_\theta$ (d), pixel-wise aligned ground truth $B_{\theta, \alpha}$ (e), synthetic image $\tilde{B}_{\theta, \alpha}$ (f), embedding$\tilde{B}^\dagger_{\theta, \alpha}$ (g).
    From top to bottom: lidar localisation against satellite imagery evaluated on KITTI (rows 1-3), radar localisation against prior lidar map evaluated on RobotCar (rows 4-6) and MulRan (7-9).
    \vspace{-2mm}
}
\end{figure*}

Additional qualitative results are presented in Figure \ref{fig:qualitative_sm1} and Figure \ref{fig:qualitative_sm2}.

\end{document}